\documentclass[runningheads]{llncs}
\usepackage{graphicx}

\usepackage{tikz}
\usepackage{comment}
\usepackage{amsmath,amssymb} 
\usepackage{color}
\newcommand{\tablesize}{{0.65}}
\usepackage{graphicx}
\usepackage{epsfig}
\usepackage[pagebackref,breaklinks,colorlinks]{hyperref}
\usepackage{bbding} 
\usepackage{bm}
\usepackage{enumerate}
\usepackage{multirow}
\usepackage{multicol}
\usepackage{colortbl}
\usepackage{color}
\usepackage{mathtools}
\usepackage{cuted}
\usepackage{capt-of}
\usepackage{subfloat}
\usepackage{subfigure}

\makeatletter
\newcommand{\@chapapp}{\relax}%
\makeatother
\usepackage[title]{appendix}

\usepackage[accsupp]{axessibility}  

\begin{document}

	\pagestyle{headings}
	\mainmatter
	\def\ECCVSubNumber{4989}  
	
	\title{RealFlow: EM-based Realistic Optical Flow Dataset Generation from Videos} 

	\titlerunning{RealFlow}
	%
	\author{Yunhui Han$^{1}$ \and
		Kunming Luo$^{2}$ \and
		Ao Luo$^{2}$ \and
		Jiangyu Liu$^{2}$ \and
		Haoqiang Fan$^{2}$ \and
		Guiming Luo$^{1}$ \and
		Shuaicheng Liu$^{3,2,\dagger}$}
	\authorrunning{Y. Han et al.}
	%
	\institute{
		School of Software, Tsinghua University,  Beijing 100084, China \\
		\email{\{hanyh19@mails.tsinghua.edu.cn, gluo@tsinghua.edu.cn\}} \\ \and  
		Megvii Technology, Beijing, China \\
		\email{\{luokunming,luoao02,liujiangyu,fhq\}@megvii.com} \\ \and  
		University of Electronic Science and Technology of China, Chengdu, China \\
		\email{liushuaicheng@uestc.edu.cn}\\  
		$^\dagger$Corresponding Author
	}
	\maketitle
	
	\begin{abstract}
		Obtaining the ground truth labels from a video is challenging since the manual annotation of pixel-wise flow labels is prohibitively expensive and laborious.
		Besides, existing approaches try to adapt the trained model on synthetic datasets to authentic videos, which inevitably suffers from domain discrepancy and hinders the performance for real-world applications.
		To solve these problems, we propose RealFlow, an Expectation-Maximization based framework that can create large-scale optical flow datasets directly from any unlabeled realistic videos. Specifically, we first estimate optical flow between a pair of video frames, and then synthesize a new image from this pair based on the predicted flow. Thus the new image pairs and their corresponding flows can be regarded as a new training set. Besides, we design a Realistic Image Pair Rendering (RIPR) module that adopts softmax splatting and bi-directional hole filling techniques to alleviate the artifacts of the image synthesis. In the E-step, RIPR renders new images to create a large quantity of training data. In the M-step, we utilize the generated training data to train an optical flow network, which can be used to estimate optical flows in the next E-step. During the iterative learning steps, the capability of the flow network is gradually improved, so is the accuracy of the flow, as well as the quality of the synthesized dataset. Experimental results show that RealFlow outperforms previous dataset generation methods by a considerably large margin. Moreover, based on the generated dataset, our approach achieves state-of-the-art performance on two standard benchmarks compared with both supervised and unsupervised optical flow methods. 
		Our code and dataset are available at \url{https://github.com/megvii-research/RealFlow}.
	\end{abstract}

	\section{Introduction}
	\label{sec:intro}
	
	\begin{figure}[t]
		\centering
		\includegraphics[width=0.7\linewidth]{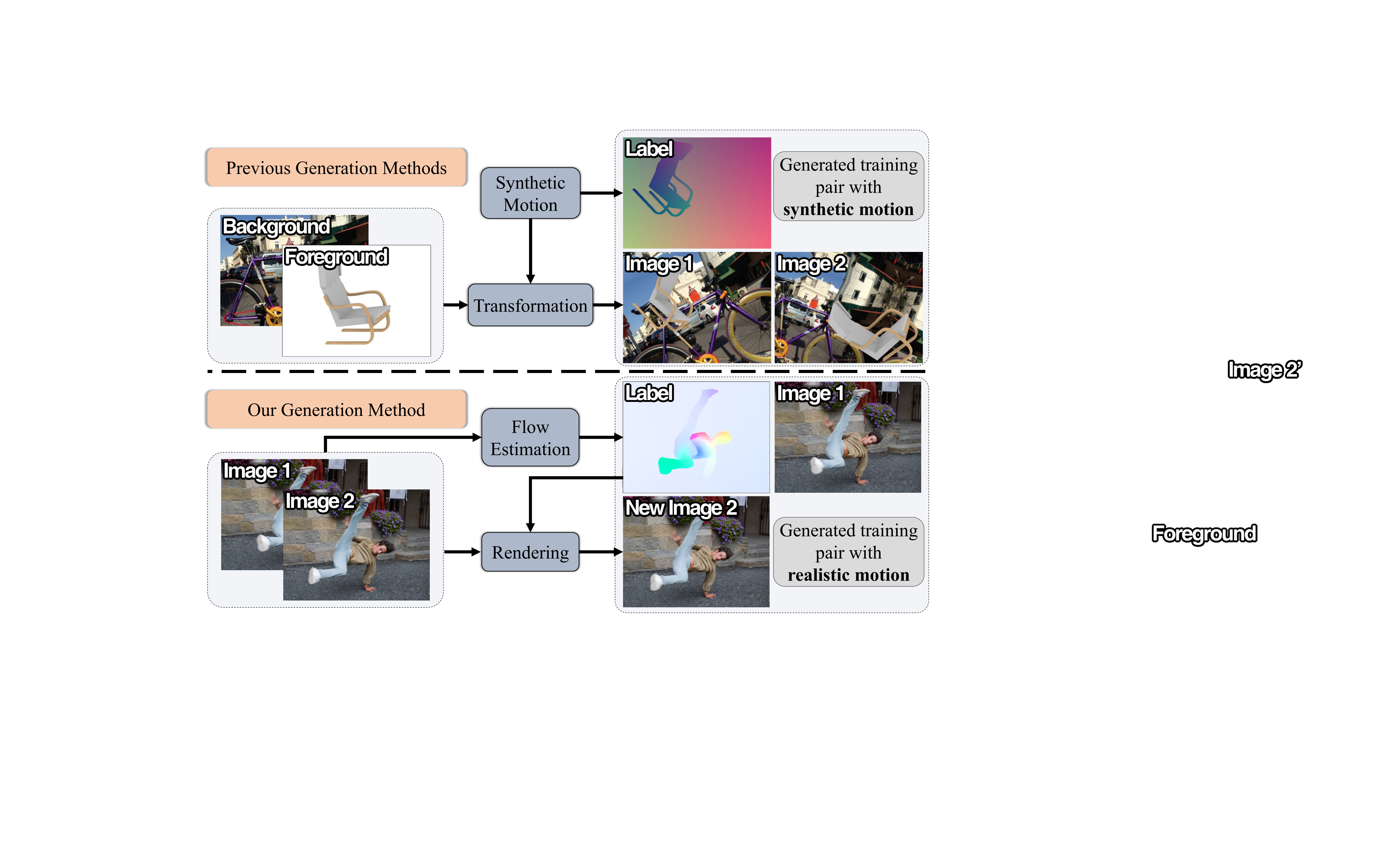}
		\caption{Illustration of our motivation. Top: previous methods use synthetic motion to produce training pairs. Bottom: we propose to construct training pairs with realistic motion labels from the real-world video sequence. We estimate optical flow between two frames as the training label and synthesize a `New Image 2'. Both the new view and flow labels are refined iteratively in the EM-based framework for mutual improvements.}\label{fig:teaser}
	\end{figure}
	
	Deep optical flow methods~\cite{raft2020,pwc_net_tpami} adopt large-scale datasets to train networks, which have achieved good computational efficiency and state-of-the-art performances in public benchmarks~\cite{KITTI_2015,butler2012naturalistic}.
	One key ingredient of these deep learning methods is the training dataset. We summarize \textbf{four} key characteristics of flow datasets that have significant impacts on the success of deep learning algorithms: \textbf{1)} the \emph{quantity} of labeled pairs; \textbf{2)} the \emph{quality} of flow labels; \textbf{3)} the \emph{image realism}; and \textbf{4)} the \emph{motion realism}. We refer to the first two as the label criteria and the latter two as the realism criteria.
	
	However, we find it is difficult for existing flow datasets to be satisfactory in all aspects.
	For example, FlyingThings~\cite{mayer2016large} synthesizes the flows by moving a foreground object on top of a background image. Sintel~\cite{butler2012naturalistic} is purely rendered from virtual 3D graphic animations. AutoFlow~\cite{sun2021autoflow} presents a learning approach searching for hyperparameters to render synthetic training pairs. As a result, these methods can produce large amounts of training data with accurate flow labels, satisfying the label criteria. However, they failed to meet the demand of realism criteria, as both the scene objects and their motions are synthesized. If flow networks are trained on these datasets, they may suffer from the domain gap between the synthetic and authentic scenes~\cite{lai2017semi}, resulting in sub-optimal performance on real-world images. 
	
	To achieve realism, some methods propose to manually annotate flow labels using realistic videos~\cite{baker2011database,liu2008human}. Although these methods can naturally satisfy the realism criteria, the process of manual labeling is time-consuming, and neither the quality nor the quantity can be guaranteed, potentially at odds with the requirements for label criteria. Recently, Aleotti~\emph{et al.}~\cite{aleotti2021learning} propose to create training pairs from a single image. It randomly generates transformations as the flow labels, based on which the image is warped to produce the other image, yielding a pair of images together with the flow labels. In this way, the label criteria are satisfied. However, the realism criteria can only be satisfied partially. Because, the synthesized images sometimes contain artifacts, and more importantly, the generated motions cannot resemble realistic object motion behaviors of real-world scenarios.
	
	\begin{figure}[t]
		\centering
		\includegraphics[width=0.7\linewidth]{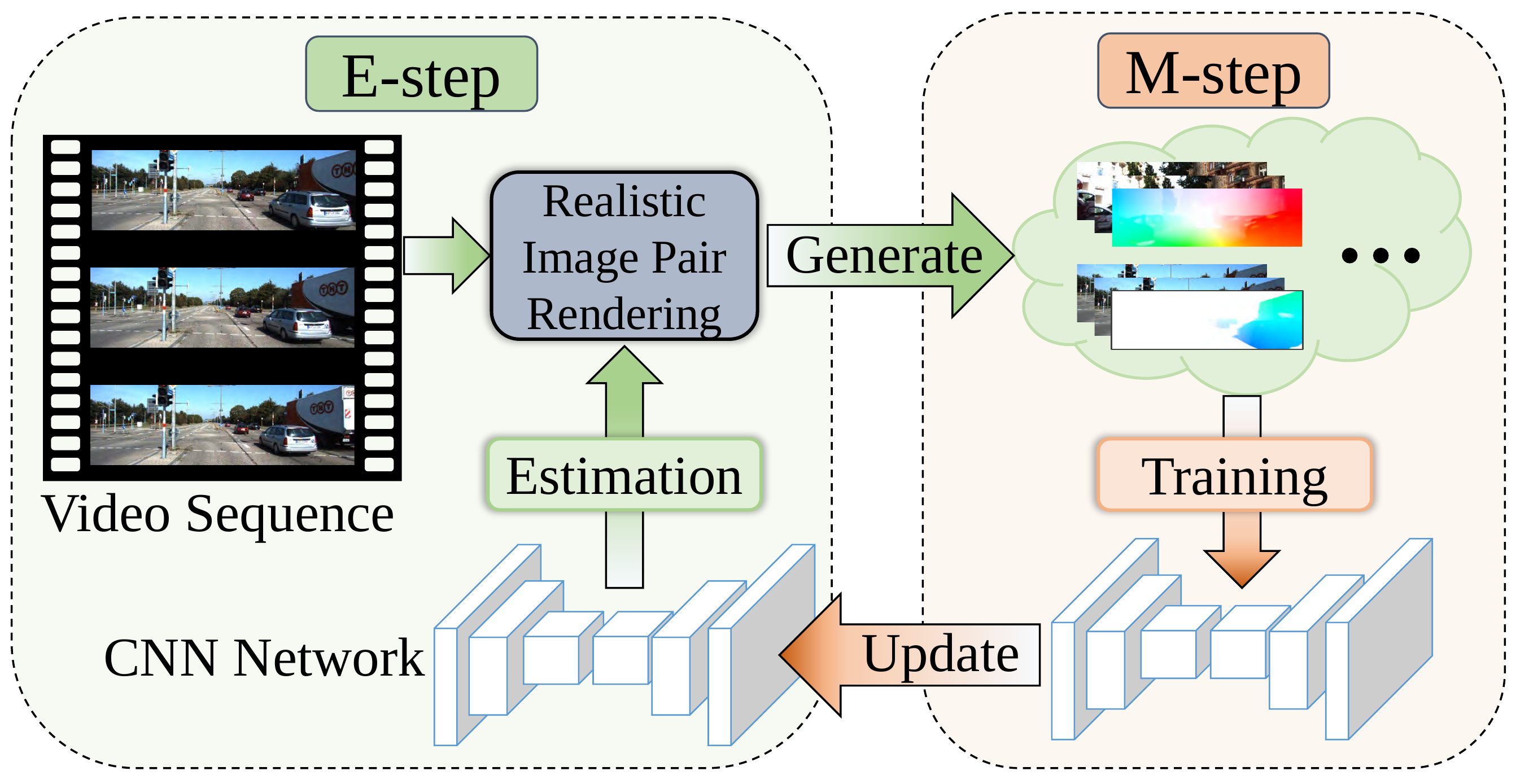}
		\caption{Illustration of the EM-based framework. In the E-step, we estimate flow labels and synthesize new views to generate training data. In the M-step, we use the training data to train a network, which can update flow labels for the next E-step.}
		\label{fig:pipeline}
	\end{figure}
	
	To address these issues, we propose RealFlow, an iterative learning framework to simultaneously generate the training pairs from realistic video frames and obtain an enhanced flow network with the generated flow data, which can satisfy both the label and realism criteria. Fig.~\ref{fig:teaser} shows an illustration of our RealFlow in comparison with existing methods. Previous work~\cite{sun2021autoflow,Flownet_flyingchairs} (Fig.~\ref{fig:teaser} top) synthesizes the flows (motions) by pasting foreground objects on top of backgrounds at different positions, where the motions are manually generated. In our approach (Fig.~\ref{fig:teaser} bottom), we first estimate the optical flows $F$ between frame pairs ($I_1$ and $I_2$) from existing videos, and then exploit the predicted flows as labels to synthesize $I'_2$, a set of new images of the `frame 2'. After that, we abandon the original $I_2$, and use the $I_1$ and $I'_2$ as image pairs, together with the estimated $F$ as the flow labels to compose a sample $(I_1, I'_2, F)$ of the new optical flow dataset. Note that the flow labels are naturally accurate for warping $I_1$ to $I'_2$, since the pixels in $I'_2$ are synthesized based on the $F$ and $I_1$.
	
	This strategy faces two challenges: 1) image synthesis may introduce artifacts, {\em e.g.,} disparity occlusions; 2) the motion realism is affected by the quality of estimated flows. 
	For the first challenge, we design Realistic Image Pair Rendering (RIPR) method to robustly render a new image $I'_2$. Specifically, we employ softmax splatting and bi-directional hole filling techniques, based on the flow and depth maps predicted from the image pairs $I_1$ and $I_2$, to generate the new images, where most of the artifacts can be effectively alleviated. 
	For the second one, we design an Expectation-Maximization (EM) based learning framework, as illustrated in Fig.~\ref{fig:pipeline}. Specifically, during the E-step, RIPR renders new images to create the training samples, and during M-step, we use the generated data to train the optical flow network that will estimate optical flows for the next E-step. During the iterative learning steps, the capability of the flow network is gradually improved, so is the accuracy of the flow, as well as the quality of synthesized dataset. Upon the convergence of our EM-based framework, we can obtain a new flow dataset generated from the input videos and a high-precision optical flow network benefiting from the new training data.
	
	By applying RealFlow, huge amounts of videos can be used to generate training datasets, which allows supervised optical flow networks to be generalized to any scene.
	In summary, our main contributions are: 
	\begin{itemize}
		\item We propose {\bf RealFLow}, an EM-based iterative refinement framework,  to effectively generate large-scale optical flow datasets with realistic scene motions and reliable flow labels from real-world videos.
		
		\item We present Realistic Image Pair Rendering ({\bf RIPR}) method for high-quality new view synthesis, overcoming issues such as occlusions and holes.
		
		\item RealFlow leads to a significant performance improvement compared against prior dataset generation methods. We generate a large real-world dataset, with which we set new records on the public benchmarks using widely-used optical flow estimation methods.
	\end{itemize}
	
	\section{Related Work}
	\noindent{\bf Supervised Optical Flow Network.} FlowNet~\cite{Flownet_flyingchairs} is the first work to estimate optical flow by training a convolutional network on synthetic dataset. 
	Following FlowNet, early approaches~\cite{ilg2017flownet,ranjan2017optical,hur2019iterative,LiteFlowNet} improve the flow accuracy with the advanced modules and network architectures. Recent works~\cite{ag2022learning,jiang2021learning,luo2022learning} propose graph and attention-based global motion refinement approaches in the recurrent framework~\cite{raft2020}, making large progress on supervised learning.
	However, for the existing supervised networks, the domain gap between synthetic datasets and realistic datasets is non-negligible and inevitably degrades the performance. Our work aims to generate datasets from realistic videos to solve this problem.
	
	\noindent{\bf Unsupervised Optical Flow Network.}
	The advantage of unsupervised methods is that no annotations are required for the training~\cite{jason2016back,ren2017unsupervised}.  Existing works \cite{Liu2019CVPR,Epipolar_flow_2019cvpr,Pengpeng2019,ren2020stflow,Li_2021_ICCV,liu2021asflow} present multiple unsupervised losses and image alignment constraints to achieve competitive results.
	However, there are many challenges for unsupervised methods, including but not limited to, occlusions~\cite{wang2018occlusion,janai2018unsupervised,liu2021oiflow}, lack of textures~\cite{simFlow2020eccv}, and illumination variations~\cite{meister2018unflow}, all of which break the basic hypothesis of brightness constancy assumption. Therefore, supervised networks achieve better performance than unsupervised ones.
	
	\noindent{\bf Dataset Generation for Optical Flow.}  Middlebury~\cite{baker2011database} records objects with fluorescent texture under UV light illumination to obtain flow labels from real-world scenes. Liu.~\emph{et al.}~\cite{liu2008human} propose a human-in-loop methodology to annotate ground-truth motion for arbitrary real-world videos. KITTI~\cite{KITTI_2012,KITTI_2015} is a popular autonomous driving dataset, which provides sophisticated training data through complex device setups. However, the quantity of the above real-world datasets is small, which is insufficient for deep supervised learning. Flyingchairs~\cite{Flownet_flyingchairs} makes the first attempt to show that synthesized training pairs can be used for supervised learning. Flyingthings~\cite{mayer2016large} further improves the quantity. Virtual KITTI~\cite{gaidon2016virtual} uses Unity game engine to create a large driving dataset. AutoFlow~\cite{sun2021autoflow} presents a learning approach to search the hyperparameter for rendering training data. However, these datasets are all synthetic. There is a constant shift from synthetic scene towards real-world scene. 
	SlowFlow~\cite{janai2017slow} attempts to collect large-scale dataset using a high-speed video camera, but the flow labels are not totally reliable.
	Efforts to tackle the above problem is Depthstillation~\cite{aleotti2021learning}, which synthesizes a new image from a single real image. The motion labels are the sampled parametric transformations for the foreground and background. However, the sampled motions are not real, and the synthesis sometimes introduces artifacts. In contrast, our method obtains reliable flow labels from real videos and synthesizes the new view from two frames instead of a single image. 
	
	\begin{figure*}[t]
		\centering
		\includegraphics[width=0.95\linewidth]{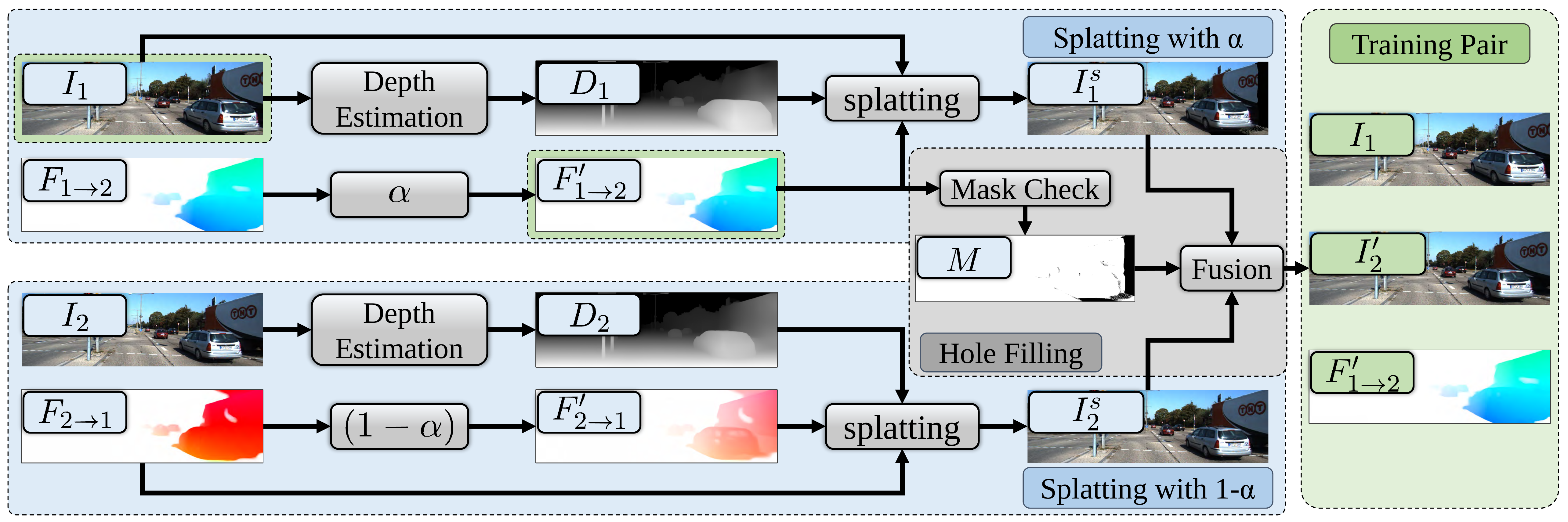}
		\caption{Illustration of Realistic Image Pair Rendering (RIPR). Given two frames $I_1$ and $I_2$ from real-world videos and their estimated flow fields $F_{1 \to 2}$ and $F_{2 \to 1}$, we first obtain the depth $D_1$ and $D_2$ by the monocular depth network\cite{ranftl2021vision}. Then, we modify the flow maps to $F_{1 \to 2}'$ and $F_{2 \to 1}'$ and use $F_{1 \to 2}'$ to check the hole mask $M$. Finally, splatting method is used to generate new views $I_{1}^s$ and $I_{2}^s$, which are further fused to render `new image 2' $I_2'$. The ($I_1$,$I_2'$,$F$) serves as the new generated training pair.}
		\label{fig:RIPR}
	\end{figure*}
	
	\section{Method}
	\label{sec:method}
	\subsection{RealFlow Framework}\label{sec:realflow_framework}
	The pipeline of the proposed RealFlow framework is illustrated in Fig.~\ref{fig:pipeline}. Given a set of real-world videos, our goal is to generate a large-scale training dataset and learn an optical flow estimation network at the same time. The key idea behind RealFlow is that a better training dataset can help learn a better optical flow network and inversely, a better network can provide better flow predictions for dataset generation. Therefore, we integrate the dataset generation procedure and optical flow network training procedure as a generative model, which can be iteratively optimized by Expectation-Maximization (EM) algorithm~\cite{mclachlan2007algorithm}. 
	\par As illustrated in Fig.~\ref{fig:pipeline}, RealFlow is an iterative framework that contains two main steps: E-step and M-step. In iteration $t$, we first conduct the E-step to generate a training dataset $X^t=\{x^t\}$. Given a consecutive image pair $(I_1, I_2)$ sampled from the input videos, the training data generation procedure can be formulated as follows:
	\begin{equation}
	x^t=\mathcal{R}(I_1, I_2,\Theta^{t-1}),
	\label{eq:E-step}
	\end{equation}
	where $\Theta^{t-1}$ is the learned optical flow network $\Theta$ in previous iteration $t-1$, $x^t$ is the generated training sample, and $\mathcal{R}$ represents our training pair rendering method Realistic Image Pair Rendering (RIPR), illustrated in Sec.~\ref{sec:RIPR}.
	\par Then, in M-step, we use the newly generated dataset $X^t$ to train and update the optical flow estimation network in a fully supervised manner:
	\begin{equation}
	\Theta^{t}=\mathop{\arg\min}\limits_{\Theta}{\mathcal{L}(X^t,\Theta)},
	\label{eq:M-step_training}
	\end{equation}
	where $\mathcal{L}$ is the learning objective of the optical flow network. 
	Finally, an optical flow dataset and a high-precision optical flow network can be obtained by RealFlow with several EM iterations. 
	
	\subsection{Realistic Image Pair Rendering}\label{sec:RIPR}
\begin{figure}[t]
		\centering
		\includegraphics[width=0.65\linewidth]{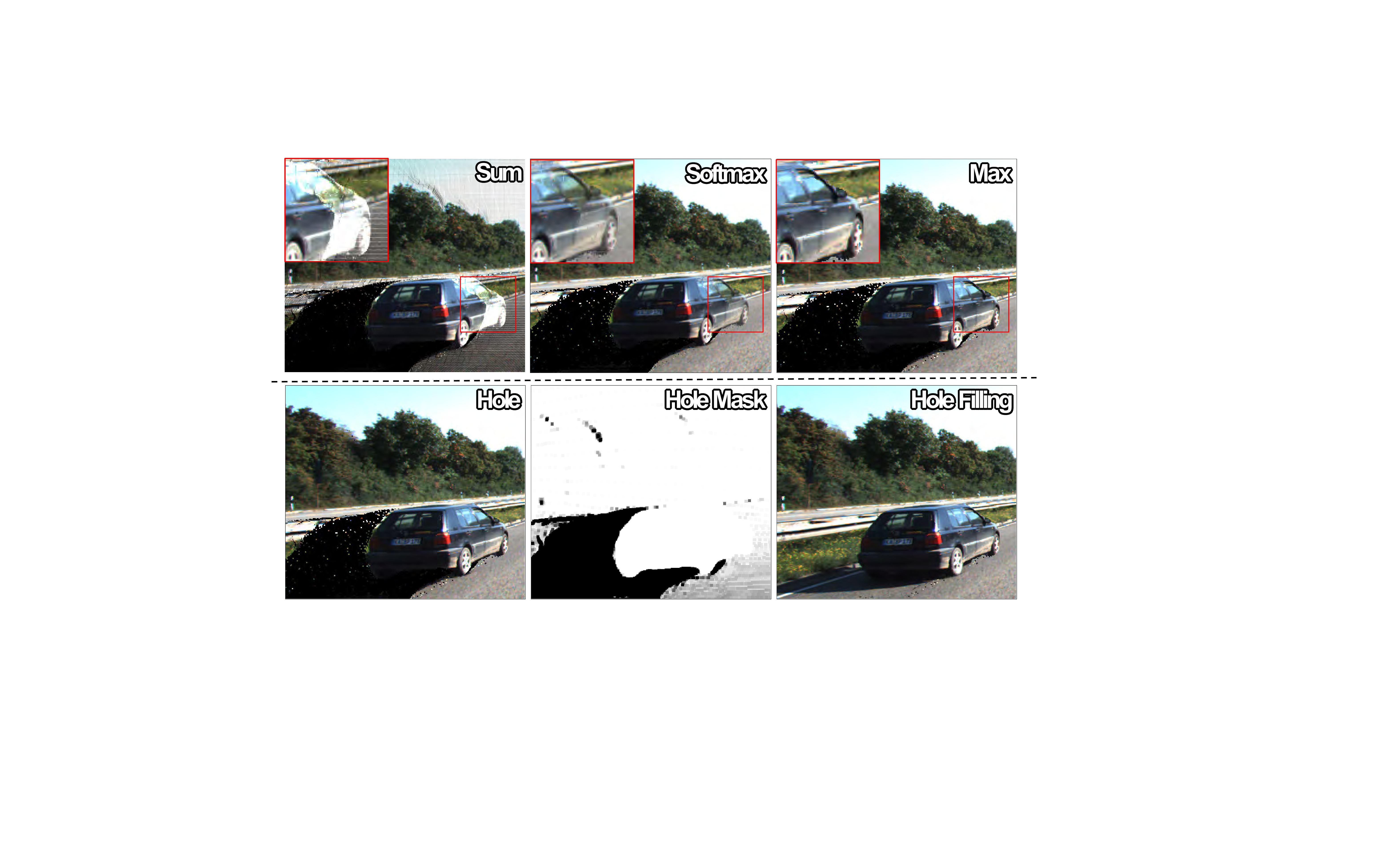}
		\caption{Examples of the splatting results (top) and the hole filling results (bottom). For the splatting, summation is a conventional approach that produces brightness inconsistency results. Softmax leads to transparent artifacts. Max splatting renders a natural image. After hole filling by our proposed BHF, a new view image with few artifacts can be generated.}
		\label{fig:splatting_and_hole_filling}
	\end{figure}

	The pipeline of the proposed RIPR method is shown in Fig.~\ref{fig:RIPR}. Given a pair of consecutive images $(I_1,I_2)$ and an optical flow estimation network $\Theta$, our goal is to generate an image pair with its flow label for network training. The main idea is to render a new image $I_2'$ based on the image pair $(I_1,I_2)$ and an estimated flow $F$ between $(I_1,I_2)$, so that $F$ can be used as the training label of the new image pair $(I_1,I_2')$. 
	Specifically, the reference image $I_1$ is first forward-warped to the target view $I_2'$. Then, in order to ensure the realism of the synthesized view $I_2'$, we need to remove the occlusions and holes caused by dynamic moving objects as well as depth disparities. Fig.~\ref{fig:splatting_and_hole_filling} illustrates an example. 
	
	Here, we use the \textbf{Splatting} method to identify foreground and background for these occlusion regions based on a monocular depth network~\cite{ranftl2021vision}.
	Moreover, we design a \textbf{Bi-directional Hole Filling (BHF)} method to fill these hole regions using backward flow and image content from $I_2$.
	Finally, after the target view generation, the reference image, synthesized new view, and the estimated flow $(I_1,I_2',F)$ are chosen as a training pair for dataset construction. 
	
	As detailed in Fig.~\ref{fig:RIPR}, we first estimate the forward flow, backward flow, and the depth of $I_1$ and $I_2$ as follows:
	\begin{align}
	F_{1 \to 2}&=\Theta(I_{1}, I_{2}),\quad & F_{2 \to 1}& =\Theta(I_{2}, I_{1}),  \label{eq:flow_estimation} \\
	D_{1}&=\Psi(I_{1}), \quad & D_{2}&=\Psi(I_{2}),  \label{eq:depth_estimation}
	\end{align}
	where $F_{1 \to 2}$ and $F_{2 \to 1}$ are the estimated forward and backward flow, and $D_{1}$ and $D_{2}$ are the estimated depth results by the monocular depth network $\Psi$. Note that $D_{1}$ and $D_{2}$ are the inverse depth maps so that the pixel with larger value is closer to the camera.
	
	In order to increase the diversity of the generated dataset, we use a factor $\alpha$ to add a disturbance to the estimated flow, so that the generated view is not exactly the original $I_2$ but a new view controlled by the factor $\alpha$. Thus we obtain new flow fields by the follows:
	\begin{equation}
	F_{1 \to 2}'=\alpha F_{1 \to 2},\quad F_{2 \to 1}'=(1-\alpha)F_{2 \to 1}.
	\label{eq:flow_disturbance}
	\end{equation}
	Then, we use flow fields $F_{1 \to 2}'$ and $F_{2 \to 1}'$ to render the new view by splatting method, which can be represented as:
	\begin{equation}
	I_{1}^s=\mathcal{S}(I_1,F_{1 \to 2}',D_1),\quad I_{2}^s=\mathcal{S}(I_2,F_{2 \to 1}',D_2),
	\label{eq:new_views_splatting}
	\end{equation}
	where $\mathcal{S}$ represents the splatting method, $I_{1}^s$ and $I_{2}^s$ are the same view rendered from different directions. Note that the occlusion problem is addressed after the splatting operation which we will introduce later. Finally, the result view can be generated by our BHF method, which is formulated as:
	\begin{equation}
	I_{2}'=\mathcal{B}(I_{1}^s,F_{1 \to 2}',I_{2}^s),
	\label{eq:BHF}
	\end{equation}
	where $\mathcal{B}$ represents our BHF method, $I_{2}'$ is the new image and $(I_1, I_{2}', F_{1 \to 2}')$ is the training pair generated of RIPR.
	
	\par {\bf Splatting.} Splatting can be used to forward-warp the reference image $I_1$ into a new view $I_s$ according to a given flow field $F_{1 \to 2}'$.
	As shown in Fig.~\ref{fig:splatting_and_hole_filling} (top), the conventional sum operation for splatting often produces brightness inconsistency results. The softmax splatting method\cite{niklaus2020softmax} is proposed to ease this problem.
	Assuming $\bm{q}$ is a coordinate in $I_1$, and $\bm{p}$ is a coordinate in the target view. The softmax splatting operation can be formulated as follows:
	\begin{align}
	\text{let }  \bm{u} &= \bm{p}-(\bm{q}+F_{1 \to 2}'(\bm{q})),  \label{eq:fwbase1} \\
	b(\bm{u}) &= \max(0,1-|\bm{u}_x|)\cdot \max(0,1-|\bm{u}_y|),  \label{eq:fwbase2}\\
	I_{s}(\bm{p}) &= \frac{\sum_{\bm{q}}{\exp{D_1(\bm{q})}\cdot I_1(\bm{q}) \cdot b(\bm{u})}}{\sum_{\bm{q}}{\exp{D_1(\bm{q})} \cdot b(\bm{u})}},  \label{eq:fwbase3}
	\end{align}
	where $b(\bm{u})$ is the bilinear kernel, $D_1$ is the depth map of $I_1$ and $I_{s}$ is the forward-warp result. By applying Eq.~\ref{eq:fwbase3}, background pixels that are occluded in the target view can be compressed by incorporating the depth map and the softmax operation, compared with the original sum splatting operation as illustrated in Fig.~\ref{fig:splatting_and_hole_filling} (top). 
	However, Softmax splatting in Eq.~\ref{eq:fwbase3} may still cause unnatural results in occlusion regions. To this end, we propose to use max splatting as an alternative option of the splatting method:
	\begin{align}
	\text{let } k&=
	\begin{cases}
	1, & \text{if }|\bm{q}+F_{1 \to 2}'(\bm{q})-\bm{p}|\leq \frac{\sqrt{2}}{2} \\
	0, & \mbox{otherwise},
	\end{cases}\label{eq:fwbase_max1} \\
	I_{s}(\bm{p}) &= I_1(\bm{r}), \text{ where } \bm{r}=\mathop{\arg\max}\limits_{\bm{q}}{D(\bm{q})\cdot k},  \label{eq:fwbase_max3}
	\end{align}
	where $k$ is the nearest kernel. 
	Eq.~\ref{eq:fwbase_max3} means that when multiple pixels are located to position $\bm{p}$, we only assign the pixel with the largest depth value to the target view. As such, the resulting image is more natural compared with the softmax version as shown in Fig.~\ref{fig:splatting_and_hole_filling} (top).
	However, we find that the dataset generated by softmax splatting performs better than the max version in our experiments. 
	Detailed analysis will be discussed in our experiment Sec.~\ref{table:Ablation experiments}.
	
	\par {\bf Bi-directional Hole Filling.} Apart from occlusions, there is another problem called holes, which are produced when no pixels from original image are projected to these regions. Previous method~\cite{aleotti2021learning} adopted an inpainting model to solve this problem, which often introduces artifacts that reduce the quality of the generated dataset. Here, we design a bi-directional hole filling method to handle these empty regions. 
	As in Eq.~\ref{eq:BHF}, the input of BHF is the forward flow $F_{1 \to 2}'$, and the target views $I_{1}^s$ and $I_{2}^s$ generated by splatting with forward and backward flows, respectively.
	We first check a hole mask $M$ from $F_{1 \to 2}'$ using the range map check method\cite{wang2018occlusion}, which is formulated as follows:
	\begin{equation}
	M(\bm{p})=\min{(1, \sum_{\bm{q}}b(\bm{u}))}, 
	\label{eq:mask_check}
	\end{equation}
	where $b(\bm{u})$ is the bilinear kernel described in Eq.~\ref{eq:fwbase2}. In the hole mask $M$, the hole pixels are labeled as 0 and others as 1.
	Then, we can generate a novel view image $I_2'$ by fusing $I_{1}^s$ and $I_{2}^s$ as follows:
	\begin{equation}
	I_2'=I_{1}^s+(1-M)\cdot I_{2}^s, 
	\label{eq:BHF_fusion}
	\end{equation}
	which means that the hole regions in $I_{1}^s$ are filled with regions in $I_{2}^s$. By applying our BHF, realistic images can be generated, which is shown in Fig.~\ref{fig:splatting_and_hole_filling} (bottom).

	\section{Experiments}
	
	\subsection{Datasets}
	
	\noindent{\bf Flying Chairs~\cite{Flownet_flyingchairs}} and {\bf Flying Things~\cite{yang2019drivingstereo}:} These two synthetic datasets are generated by randomly moving foreground objects on top of a background image. State-of-the-art supervised networks usually train on Chairs and Things.
	
	\noindent{\bf Virtual KITTI~\cite{gaidon2016virtual}:} Virtual KITTI is a synthetic dataset, which contains videos generated from different virtual urban environments.
	
	\noindent{\bf DAVIS~\cite{Caelles_arXiv_2019}:} DVAIS dataset consists of high-quality video sequences under various kinds of scenes. No optical flow label is provided. We use 10,581 images from DAVIS challenge 2019 to generate {\bf RF-DAVIS}.
	
	\noindent{\bf ALOV~\cite{smeulders2013visual} and BDD100K~\cite{yu2020bdd100k}:} 
	ALOV and BDD100K datasets are large-scale real-world video databases. We capture 75,581 image pairs from ALOV dataset and 86,128 image pairs from BDD100K dataset. There is no flow label for these image pairs, so we use RealFlow to create a large diverse real-world dataset with flow label, named {\bf RF-AB}.
	
	\noindent{\bf KITTI~\cite{KITTI_2012,KITTI_2015}:} KITTI2012 and KITTI2015 are benchmarks for optical flow estimation. There are multi-view extensions (4,000 training and 3,989 testing) datasets with no ground truth. We use the multi-view extension videos (training and testing) of KITTI 2015 to generate {\bf RF-Ktrain} and {\bf RF-Ktest} datasets.
	
	\noindent{\bf Sintel~\cite{butler2012naturalistic}:} Sintel is a synthetic flow benchmark derived from 3D animated film, which contains 1,041 training pairs and 564 test pairs. We use the images from the training set to generate our {\bf RF-Sintel}.
	
	\begin{figure*}[t]
		\centering
		\includegraphics[width=0.9\linewidth]{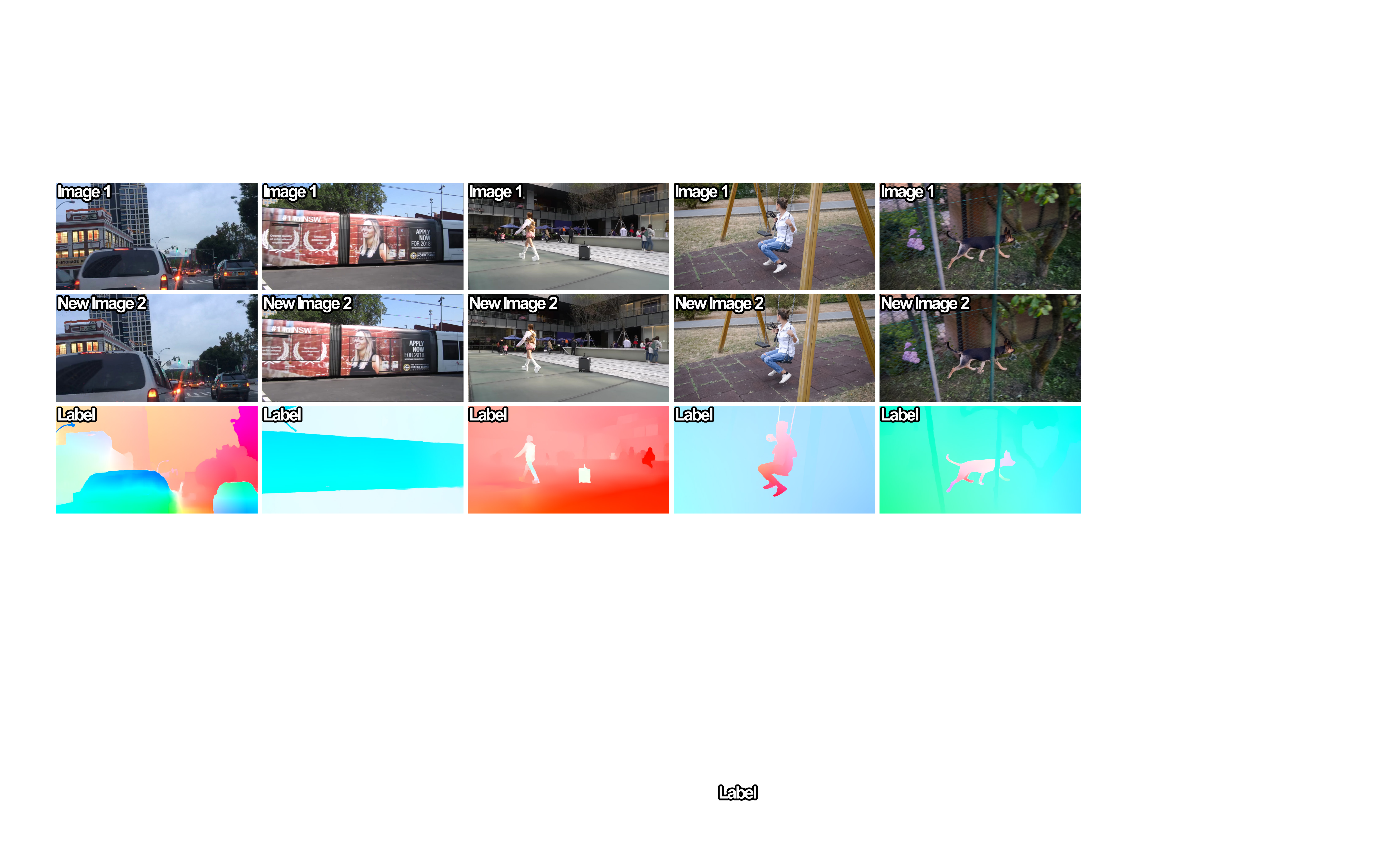}
		\caption{Example training pairs from our generated RF-AB and RF-DAVIS. The first sample contains large motion and complex scenes.}
		\label{fig:results_dataset_sample}
	\end{figure*}
	
	\subsection{Implementation Details}
	
	Our RIPR consists of a depth estimation module, a flow estimation module, a splatting module, and a hole filling module. For the flow estimation module, we select RAFT~\cite{raft2020} which represents state-of-the-art architecture for supervised optical flow. We train the RAFT using official implementation without any modifications. We initialized the RealFlow framework using RAFT pre-trained on FlyingChairs and FlyingThings unless otherwise specified. For the depth estimation module, we select DPT~\cite{ranftl2021vision} monocular depth network, which represents the state-of-the-art architecture. For the splatting module, softmax splatting~\cite{niklaus2020softmax} is used due to the better performance. For the hole filling, our BHF uses the bi-directional flow estimated from RAFT. We will show the performance of our RIPR method affected by the different settings of above modules in Sec.~\ref{sec:Ablation Study}. 
	
	\subsection{Comparison with Existing Methods}
	
	In this section, we evaluate the effectiveness of RealFlow generation framework on the public benchmarks.
	
	\noindent{\bf Comparison with Dataset Generation Methods.} 
	Due to the scarcity of real-world dataset generation methods for optical flow, we only select the Depthstillation method~\cite{aleotti2021learning} for comparison. 
	Depthstillation generated optical flow dataset dDAVIS and dKITTI from DAVIS and KITTI multi-view test. For fair comparison, we also choose the DAVIS and KITTI multi-view test videos to generate our RF-DAVIS and RF-Ktest. 
	Fig.~\ref{fig:results_dataset_sample} shows our rendered training pairs.
	We evaluate our method on KITTI12-training and KITTI15-training sets. Quantitative results are shown in Table~\ref{table:comparision_with_generation_method}, where our method outperforms Depthstillation~\cite{aleotti2021learning}, proving the importance of realism of object motion behavior.

	\begin{table}[t]
		\centering
		\caption{Comparison with previous dataset generation method\cite{aleotti2021learning}. We use the same source images to generate dataset and train the same network for comparison. The best results are marked in {\color{red}{red}}.}
		\resizebox*{\tablesize \linewidth}{!}{
			\begin{tabular}
				{
					>{\arraybackslash}p{1.0cm} 
					>{\centering\arraybackslash}p{2.0cm}| 
					>{\centering\arraybackslash}p{1.2cm} 
					>{\centering\arraybackslash}p{1.2cm}| 
					>{\centering\arraybackslash}p{1.2cm} 
					>{\centering\arraybackslash}p{1.2cm} 
				}
				\hline
				\multirow{2}{*}{Model}  & \multirow{2}{*}{Dataset} & \multicolumn{2}{c|}{KITTI12} & \multicolumn{2}{c}{KITTI15} \\
				&& EPE     & F1    & EPE & F1       \\
				\hline
				RAFT  &   dDAVIS\cite{aleotti2021learning}   & 1.78   &6.85\%  &3.80    & 13.22\%   \\
				RAFT  &   RF-DAVIS                           & \textcolor{red}{1.64}   &\textcolor{red}{5.91\%}   &\textcolor{red}{3.54}    & \textcolor{red}{9.23\%}    \\
				\hline
				RAFT  &   dKITTI\cite{aleotti2021learning}     & 1.76   &5.91\%     &4.01   & 13.35\%    \\
				RAFT  &   RF-Ktest  & \textcolor{red}{1.32}   &\textcolor{red}{5.41\%}    &\textcolor{red}{2.31}   &\textcolor{red}{8.65\%}   \\
				\hline
		\end{tabular}}
		\label{table:comparision_with_generation_method}
	\end{table}
	
	\begin{table}[t]
		\centering
		\caption{Comparison with Unsupervised Methods.The best results are marked in {\color{red}{red}} and the second best are in {\color{blue}{blue}}, `-' indicates no results. End-point error (epe) is used as the evaluation metric. }
		\resizebox*{0.7 \linewidth}{!}{
			\begin{tabular}{
					>{\arraybackslash}p{3.8cm}|
					>{\centering\arraybackslash}p{1.3cm}| 
					>{\centering\arraybackslash}p{1.3cm}| 
					>{\centering\arraybackslash}p{1.3cm}| 
					>{\centering\arraybackslash}p{1.3cm} 
				}
				\hline
				{Method} & {KITTI12} & {KITTI15} &{Sintel C.} &{Sintel F.}\\
				\hline
				ARFlow~\cite{liu2020learning}              &1.44  &2.85  &2.79  &3.87 \\
				SimFlow~\cite{simFlow2020eccv}             & --   &5.19  &2.86  &3.57 \\
				UFlow~\cite{jonschkowski2020matters}       &1.68  &2.71  &2.50  &3.39 \\
				UpFlow~\cite{luo2021upflow}                & \textcolor{blue}{1.27} & 2.45 &2.33  &2.67 \\
				SMURF~\cite{stone2021smurf}                & --  &\textcolor{red}{2.00} &1.71 &\textcolor{blue}{2.58} \\
				\hline
				IRRPWC(C+T)~\cite{hur2019iterative} & 3.49  & 10.21  & 1.87  & 3.39  \\
				IRRPWC(C+T)+UpFlow                  & 1.87  & 2.62  & 1.79  & 3.31   \\
				Ours(IRRPWC)                         & 1.83  & 2.39  & 1.74  & 3.20  \\
				\hline
				RAFT(C+T)\cite{raft2020}                                 &2.15  &5.04  &\textcolor{blue}{1.43} &2.71\\
				Ours(RAFT)                                       & \textcolor{red}{1.20}   &\textcolor{blue}{2.16}  &\textcolor{red}{1.34} &\textcolor{red}{2.38}\\
				\hline
			\end{tabular}
		}
		\label{table:comparision_with_unsupervised_method}
	\end{table}
	
	\noindent{\bf Comparison with Unsupervised Methods.}
	When supervised optical flow networks are trained on synthetic datasets, they are hard to be generalized to real-world data due to the domain gap and motion discrepancy between synthetic and authentic datasets. 
	To some extent, the effectiveness of our method depends on domain adaptation. 
	Given the rich literature of unsupervised methods, we compare our method with them to exclude the influence of the domain. We train the RAFT~\cite{raft2020} on RF-Ktrain and RF-Sintel by RealFlow framework. 
	As shown in Table~\ref{table:comparision_with_unsupervised_method}, RealFlow outperforms all the unsupervised methods on Sintel-training. We obtain a competitive result on KITTI15-training which surpass all the unsupervised methods except SMURF\cite{stone2021smurf}. One reason is that SMURF adopted multiple frames for training, while RealFlow only uses two frames.
	
	Since our method is based on a model pre-trained on C+T (FlyChairs and FlyThings) in a supervised manner, we also provide the results of unsupervised methods that are pre-trained with groundtruth on C+T for a fair comparison. Because we cannot implement SMURF, we use IRRPWC~\cite{hur2019iterative} structure and UpFlow~\cite{luo2021upflow}  for comparison. 
	Specifically, we use IRRPWC pre-trained on C+T as a baseline, which is `IRRPWC(C+T)' in Table~\ref{table:comparision_with_unsupervised_method}. Then we train IRRPWC from the C+T pre-trained weights using unsupervised protocol provided by UpFlow on KITTI 2015 multi-view videos and Sintel sequences and do evaluation on KITTI 2012/2015 train and Sintel train data sets, which is `IRRPWC(C+T)+UpFlow'. Finally, we perform our method using IRRPWC(C+T), which is `Ours(IRRPWC)'. As a result, Our method can achieve better performance than unsupervised method trained from C+T pre-trained weights.

	\noindent{\bf Comparison with Supervised Methods.}
	To further prove the effectiveness of RealFlow, we use KITTI15-training to fine-tune the RAFT model pre-trained by our RF-Ktrain. Note that RF-Ktrain is generated without any sequence that contains the frames in KITTI test set. The evaluation results on KITTI15-training and KITTI15-testing are shown in Table~\ref{table:comparision_with_supervised_method}. We achieve state-of-art performance on KITTI 2015 test benchmark compared to previous supervised methods.
	
	\begin{table}[t]
		\centering
		\caption{Comparison of our method with supervised methods on KITTI2015 train set and test set. `-' indicates no results reported. }
		\resizebox*{0.6 \linewidth}{!}{
			\begin{tabular}
				{
					>{\arraybackslash}p{2.4cm}| 
					>{\centering\arraybackslash}p{1.4cm} 
					>{\centering\arraybackslash}p{1.4cm}| 
					>{\centering\arraybackslash}p{2.0cm} 
				}
				\hline
				\multirow{2}{*}{method}  & \multicolumn{2}{c|}{KITTI15(train)} & KITTI15(test) \\
				& EPE     & F1    & F1       \\
				\hline
				PWC-Net\cite{pwc_net}              & 2.16   &9.80\%    & 9.60\%  \\
				LiteFlowNet\cite{LiteFlowNet}      & 1.62   &5.58\%    & 9.38\%  \\
				IRR-PWC\cite{hur2019iterative}     & 1.63   &5.32\%    & 7.65\%  \\
				RAFT\cite{raft2020}                & 0.63   &1.50\%    & 5.10\%  \\
				RAFT-RVC\cite{sun2020tf}           & -   &-    & 5.56\%  \\
				AutoFlow\cite{sun2021autoflow}     & -  & -     & 4.78\%  \\
				Ours         & \textcolor{red}{0.58} & \textcolor{red}{1.35\%}   & \textcolor{red}{4.63\%} \\
				\hline
		\end{tabular}}
		\label{table:comparision_with_supervised_method}
	\end{table}

	\noindent{\bf Comparison on Large Datasets.} To make our trained networks general, we collect a large-scale realistic dataset named RF-AB. We train the RAFT from scratch using our RF-AB as official implementation. Because of the scarcity of real-world evaluation benchmarks, we only evaluate our dataset on KITTI and Sintel. As summarized in Table~\ref{table:comparision_with_large_datasets}, RAFT trained on RF-AB is more accurate than on other datasets when evaluated on KITTI12-training and KITTI15-training, which demonstrates the generalization ability of our method on real-world scenes. We also obtain comparable results on Sintel, which only surpass dCOCO~\cite{aleotti2021learning}. 
	RF-AB and dCOCO are both real-world datasets. The networks trained on them are hardly adapted to Sintel (synthetic data). So their performance is worse than C+T and Autoflow.
	Moreover, AutoFlow~\cite{sun2021autoflow} learns the hyperparameters to render training data using the average end-point error (AEPE) on Sintel as the learning metric. FlyChairs and FlyingThings are also rendered to match the displacement distribution of Sintel. Mayer~\emph{et al.}~\cite{mayer2018makes} shows that matching the displacement statistics of the test data is important.
	
	\begin{table}[t]
		\centering
		\caption{Comparison with large datasets. `-' indicates no results. End-point error (epe) is used as the evaluation metric.}
		\resizebox*{\tablesize \linewidth}{!}{
			\begin{tabular}
				{
					>{\arraybackslash}p{0.8cm} 
					>{\centering\arraybackslash}p{2.1cm}| 
					>{\centering\arraybackslash}p{1.4cm} 
					>{\centering\arraybackslash}p{1.4cm} 
					>{\centering\arraybackslash}p{1.4cm} 
					>{\centering\arraybackslash}p{1.4cm} 
				}
				\hline
				{Model}  & {Dataset} & {KITTI12} & {KITTI15} & {Sintel C.} & {Sintel F.}\\
				\hline
				RAFT  & C+T\cite{raft2020}    & 2.15     & 5.04       & \textcolor{red}{1.43}          & \textcolor{blue}{2.71}                      \\
				RAFT  & AutoFlow\cite{sun2021autoflow} & --                          & 4.23                        & 1.95                         &\textcolor{red}{2.57}                         \\
				RAFT  & dCOCO\cite{aleotti2021learning}    & \textcolor{blue}{1.82}                        & \textcolor{blue}{3.81}                        & 2.63                          & 3.90                       \\\hline
				RAFT  & RF-AB    & \textcolor{red}{1.80}                        & \textcolor{red}{3.48}                        & \textcolor{blue}{1.80}                          & 3.28
				\\ \hline
		\end{tabular}}
		\label{table:comparision_with_large_datasets}
	\end{table}

	\noindent{\bf Impact on Different Optical Flow Networks.}
	In Table~\ref{table:different networks}, we also provide experiment results to prove that our method can improve other supervised networks on real-world scenes not only on specific architecture such as RAFT. For fair comparison, we trained IRR-PWC~\cite{hur2019iterative} and GMA~\cite{jiang2021learning} on RF-AB and RF-Sintel with the official settings. Table~\ref{table:different networks} shows that RAFT and GMA trained on RF-AB outperform the original variants trained on C+T when testing on real-world data KITTI. Moreover, there is a significant improvement on IRR-PWC which is effective as trained RAFT on C+T. This fact proves that a better dataset is crucial to a supervised network.

	\subsection{Ablation Study}\label{sec:Ablation Study}
	In this section, we conduct a series of ablation studies to analyze the impact of different module choices of the RIPR method. We measure all the factors using RF-Ktrain to train RAFT and evaluate on KITTI12-training and KITTI15-training. Because there are multiple combinations of these factors, we only test a specific component of our approach in isolation. As shown in Table~\ref{table:Ablation experiments}, default settings are underlined and detail experiment settings will be discussed below. 
	
	\noindent{\bf Render.}
	We conduct an experiment named `Render Off' where we use original image pairs and their estimated flows to train the network. When applying our RIPR method, as the `Render On' in Table~\ref{table:Ablation experiments}, the accuracy of the network can be improved significantly. Moreover, our rendering method is also related to the video interpolation methods~\cite{xu2019quadratic,huang2022rife}. We replace our rendering method with QVI~\cite{xu2019quadratic} for fair comparison. Note that `QVI(RAFT)' uses RAFT pre-trained on C+T for optical flow estimation, which is the same model as the initial model of RealFlow. As a result, our method outperforms QVI for optical flow dataset generation because the frame synthesis process in QVI may cause the content of the generated frame to not match the optical flow label.
	
	\noindent{\bf Depth.}
	To measure the effectiveness of the depth estimation in the splatting method, we conduct three different experiments: DPT\cite{ranftl2021vision}, Midas\cite{ranftl2020towards}, and `Occ-bi'. The `Occ-bi' means that the depth map is replaced by the occlusion map produced by the bi-directional flow check method\cite{meister2018unflow}. DPT is a state-of-art method that outperforms Midas in the task of monocular depth estimation. From Table~\ref{table:Ablation experiments}, we can notice that with more accurate depth estimation results, our RealFlow can generate better dataset for optical flow learning, which proves that depth is a crucial cue in our framework.
	
	\begin{table}[t]
		\centering
		\caption{Impact on different optical flow networks. The value in the bracket means the percentage of improvement.End-point error (epe) is used as the evaluation metric.}
		\resizebox*{\tablesize \linewidth}{!}{
			\begin{tabular}{lc|cccc}
				\hline
				Model   & Dataset & \multicolumn{1}{l}{KITTI12} & \multicolumn{1}{l}{KITTI15} & \multicolumn{1}{l}{Sintel C.} & \multicolumn{1}{l}{Sintel F.} \\\hline
				IRR-PWC\cite{hur2019iterative}  & C+T     & 3.49                        & 10.21                       & 1.87                          & 3.39                          \\
				IRR-PWC & RF-AB   & \textcolor{red}{2.13}                        & \textcolor{red}{5.09}                           & 3.68                          & 4.72                          \\
				IRR-PWC  & RF-Sintel     &2.67                        &7.06                       &\textcolor{red}{1.74}                         &\textcolor{red}{3.20}                               \\\hline
				GMA\cite{jiang2021learning}   & C+T     & 1.99                        & 4.69                        & 1.30                          & 2.73                          \\
				GMA     & RF-AB   & \textcolor{red}{1.82}                         & \textcolor{red}{3.64}                         & 1.93                          & 3.45                          \\
				GMA   & RF-Sintel  &1.74                         &4.39                       &\textcolor{red}{1.23}                  &\textcolor{red}{2.32}      \\\hline
				RAFT\cite{raft2020}     & C+T     & 2.15                        & 5.04                        & 1.43                          & 2.71                          \\
				RAFT    & RF-AB   & \textcolor{red}{1.80}                        & \textcolor{red}{3.48}                         & 1.80                          & 3.28                          
				\\ 
				RAFT    & RF-Sintel      &1.76                        &4.36                      &\textcolor{red}{1.34}                         &\textcolor{red}{2.38}                         \\\hline
			\end{tabular}
		}
		\label{table:different networks}
	\end{table}
	
	\begin{table}[t]
		\centering
		\caption{Ablation experiments. Settings used in our final framework are underlined. Here we only perform one EM iteration for these experiments due to the limitation of computational resources.}
		\resizebox*{\tablesize \linewidth}{!}{
			\begin{tabular}
				{
					>{\arraybackslash}p{1.3cm} 
					>{\centering\arraybackslash}p{2.4cm}| 
					>{\centering\arraybackslash}p{1.0cm} 
					>{\centering\arraybackslash}p{1.0cm} 
					>{\centering\arraybackslash}p{1.0cm} 
					>{\centering\arraybackslash}p{1.0cm} 
				}
				\hline
				\multirow{2}{*}{Experiment}  & \multirow{2}{*}{Method} & \multicolumn{2}{c}{KITTI12} & \multicolumn{2}{c}{KITTI15} \\
				& &EPE &F1&EPE &F1\\
				\hline
				\multirow{4}{*}{Render} &Off &1.80&7.78\% & 3.93 &14.38\% \\
				&QVI~\cite{xu2019quadratic} &2.84&10.7\% & 7.27 &20.36\% \\
				&QVI(RAFT) &4.03&14.0\% & 9.03 &24.70\% \\
				&\underline{On}&1.44&5.90\% & 2.79 & 10.66\% \\
				\hline
				\multirow{3}{*}{Depth} & Occ-bi & 1.51 & 6.22\% & 3.01 & 11.12\%\\
				&MiDas\cite{ranftl2020towards} & 1.49 & 6.25\% & 2.90 & 11.18\% \\
				& \underline{DPT}\cite{ranftl2021vision} & 1.44 & 5.90\% & 2.79 & 10.66\%\\
				\hline
				\multirow{2}{*}{Splatting}&Max & 1.62 & 5.90\% & 3.03 & 11.04\%\\
				&\underline{Softmax}\cite{niklaus2020softmax}&1.44 & 5.90\% & 2.79 & 10.66\%\\
				\hline
				\multirow{3}{*}{Hole Filling} &w/o Filling & 1.45& 6.06\%& 2.95& 10.80\%\\
				&RFR\cite{li2020recurrent}&1.53& 6.07\%& 2.95& 11.23\%\\
				&\underline{BHF}& 1.44& 5.90\%&2.79& 10.66\%\\			
				\hline
				\multirow{3}{*}{Range of $\alpha$} &{[} 1 {]}&1.57 &6.34\%& 3.38& 12.30\%\\
				&{[}-2,2{]}&1.45 &5.92\%&2.83 &10.90\%\\
				&\underline{{[}0,2{]}}&1.44 &5.90\%& 2.79 &10.66\%\\
				\hline
		\end{tabular}}
		\label{table:Ablation experiments}
	\end{table}
	
	\begin{table}[t]
		\centering
		\caption{Iteration times. The best iteration time is underlined. Our RealFlow can converge to similar results with different initialization settings.}
		\resizebox*{\tablesize \linewidth}{!}{
			\begin{tabular}
				{ 
					>{\arraybackslash}p{1.2cm}
					>{\centering\arraybackslash}p{1.2cm} 
					>{\centering\arraybackslash}p{1.6cm}| 
					>{\centering\arraybackslash}p{1.2cm} 
					>{\centering\arraybackslash}p{1.2cm} 
					>{\centering\arraybackslash}p{1.2cm} 
					>{\centering\arraybackslash}p{1.2cm} 
				}
				\hline
				\multirow{2}{*}{Model} & {Initialize}  & {Iteration} & \multicolumn{2}{c}{KITTI12} & \multicolumn{2}{c}{KITTI15} \\
				&Dataset&Times &EPE &F1&EPE &F1\\
				\hline
				\multirow{6}{*}{RAFT} & \multirow{6}{*}{C+T} & init & 2.15 & 9.29\%  & 5.04 & 17.4\%  \\
				&&Iter.1 &1.44 & 5.90\% & 2.79 & 10.7\%  \\
				&&Iter.1$\ast 4$&1.45 & 5.59\% & 2.86 & 10.4\%  \\
				&&Iter.2 &1.31 & 5.28\% & 2.36 & 8.46\%  \\
				&&Iter.3 &1.28 & 5.02\%  & 2.20 & 8.27\%  \\
				&&\underline{Iter.4} & 1.27 & 5.17\% & 2.16 & 8.45\%\\
				\hline
				\multirow{2}{*}{RAFT}&\multirow{2}{*}{VKITTI} &init&1.81 & 5.04\% & 3.13 & 8.63\%\\
				&&\underline{Iter.1}&1.26 & 4.47\%  & 2.11 & 7.50\% \\
				\hline
				\multirow{2}{*}{GMA} & \multirow{2}{*}{C+T} & init & 1.99 & 9.28\%  & 4.69 & 17.1\%  \\
				&&\underline{Iter.1}&1.46 & 5.56\% & 2.79 & 10.2\%  \\
				\hline
		\end{tabular}}
		\label{table:Iteration Times}
	\end{table}
	
	\noindent{\bf Splatting.}
	We compared two versions of splatting: Max and Softmax\cite{niklaus2020softmax}. 
	Max splatting leads to the right rendering result of visual appearance.
	However, we find that Softmax splatting outperforms Max splatting as in Table~\ref{table:Ablation experiments}. 
	The reason is that max splatting may cause tearing of texture when the depth is incorrect, while softmax splatting can alleviate this problem by generating a translucent fusion result. Please refer to supplementary materials for more details.
	
	\noindent{\bf Hole Filling.}
	The optical flow network learns a per-pixel matching of two images. The hole in the newly generated image means that there is no pixel matched to the reference image. Although it happens, the context information can also help the network. For fair comparison, we use the RFR\cite{li2020recurrent} fine-tuned on KITTI dataset for inpainting these holes. As summarized in Table~\ref{table:Ablation experiments}, our designed BHF method achieves the best results. We also conduct an experiment without hole filling which leads to a moderate improvement over RFR. It suggests that a worse hole filling result may introduce negative effects.
	
	\noindent{\bf Range of $\alpha$.}
	To increase the diversity of our generated dataset, we add a disturbance to our RealFlow by $\alpha$, which is introduced in Sec.~\ref{sec:realflow_framework}. We use three different settings in Table~\ref{table:Ablation experiments} `Range of $\alpha$'. `$[1]$' means that $\alpha$ is always set as $1$. The other two settings mean that we randomly sample a value within that range. As can be seen, factor $\alpha$ sampled from range $[0,2]$ achieves better result.
	
	\noindent{\bf EM Iteration Times.} In RealFlow framework, the generated dataset and the optical flow network are gradually improved after iterations. However, a certain upper limit exists in RealFlow and it will converge after several iterations. As summarized in Table~\ref{table:Iteration Times}, after 4 iterations, RealFlow converges and the result cannot be further improved. 
	`Iter.1$\ast 4$' means that the network is trained $4$ times longer (more training steps) with the data of `Iter.1'. As can be seen, simply training $4$ times longer cannot bring improvement compared with $4$ EM iterations of RealFlow (see `Iter.4'), which demonstrates the effectiveness of our approach.
	
	\noindent{\bf Initial Model.} It is well-known that the initialization is important for EM algorithm. In Table~\ref{table:Iteration Times}, we implement RAFT pre-trained on Virtual KITTI (VKITTI) and GMA pre-trained on C+T as the initial model of RealFlow. As can be seen, the performance can be improved after learning with our RealFlow.
	
	\section{Conclusions}
	
	In this work, we have presented RealFlow, an EM-based framework for optical flow dataset generation on realistic videos. We have proposed a Realistic Image Pair Rendering (RIPR) method to render a new view from a pair of images, according to the estimated optical flow. We have trained optical flow networks on the synthesized dataset. Experiment results show that the trained networks and the generated datasets can be improved iteratively, yielding a large-scale high-quality flow dataset as well as a high-precision optical flow network.
	Experiments show that the performance of existing methods can be largely improved on widely-used benchmarks while using our RealFlow dataset for training.
	
	\noindent{\bf Acknowledgement}: This work was supported by the National Natural Science Foundation of China (NSFC) No.62173203, No.61872067 and No.61720106004.

	%
	%

	\newpage
	\begin{appendices}
		\chapter*{Appendix}
		To make our RealFlow self-contained, we provide more details in this document including: 1) detailed information of the datasets 2) experiments of the dataset quantity. 3) results of splatting and hole filling. 4) the efficiency of generation. 5) limitation of our method. 6) more qualitative comparison results and visualization of RealFlow samples from various videos.

		\begin{table}[t]
			\centering
			\caption{An overview of our created datasets.}
			\resizebox*{\tablesize \linewidth}{!}{
				\begin{tabular}{l|c|l}
					\hline
					Dataset   & Quantity & \multicolumn{1}{c}{Source Data}                                                           \\ \hline
					RF-Ktrain & 4,000    & KITTI-multiview(training)       \\\hline
					RF-Ktest  & 3,989    &  KITTI-multiview(testing)        \\\hline
					RF-Sintel & 1,041    &  Sintel-Clean(training)       \\ \hline
					RF-DAVIS  & 10,581   &  DAVIS-2019 video sequence    \\\hline
					RF-AB     & 161,709  &  ALOV and  BDD100K video sequence  \\\hline
				\end{tabular}
			}
			\label{table:dataset_discribe}
		\end{table}
		
		\begin{table}[t]
			\centering
			\caption{Analyze the impact of the dataset quantity. We use the datasets with different amounts of training pairs to train RAFT with the same implementation. The learned models are evaluated on KITTI 2015 train set and Sintel Final train set, where the end-point error (epe) is used as the evaluation metric.}
			\resizebox*{0.5 \linewidth}{!}{
				\begin{tabular}
					{
						>{\arraybackslash}p{1.8cm} 
						>{\centering\arraybackslash}p{1.3cm}| 
						>{\centering\arraybackslash}p{1.3cm} 
						>{\centering\arraybackslash}p{1.3cm} 
					}
					\hline
					{Dataset}  & {Quantity} & {KITTI15} & {Sintel F.}\\
					\hline
					dCOCO\cite{aleotti2021learning}  & 100k  &3.81     &3.90 \\ 
					RF-AB  & 10k   &3.83     &3.51 \\
					RF-AB  & 40k   &3.65     &3.35 \\
					RF-AB  & 80k   &3.60     &3.32 \\
					RF-AB  & 160k  &3.48     &3.28 \\
					\hline
			\end{tabular}}
			\label{table:amount}
		\end{table}
		
		\section{Datasets Detail}
		In this section, we describe the detailed information of relevant datasets and the generated datasets of our RealFlow. For convenience, we summarize our generated datasets in Table.\ref{table:dataset_discribe}. 
		
		\noindent{\bf Flying Chairs~\cite{Flownet_flyingchairs}} and {\bf Flying Things~\cite{yang2019drivingstereo}:} These two synthetic datasets are generated by randomly moving foreground objects on top of a background image. State-of-the-art supervised networks usually train on Chairs and Things.
		
		\noindent{\bf Virtual KITTI~\cite{gaidon2016virtual}:} Virtual KITTI is a synthetic dataset, which contains 50 high-resolution videos generated from $5$ different virtual urban environments under different weather conditions. These environments are created by the Unity engine and the flow labels are automatically generated.
		
		\noindent{\bf DAVIS~\cite{Caelles_arXiv_2019}:} DVAIS dataset consists of high-quality video sequences under various kinds of scenes, which have been widely used for video object segmentation. It only provides densely annotated segmentation labels but not optical flow labels. We use 10,581 images from DAVIS challenge 2019 to generate {\bf RF-DAVIS}.
		
		\noindent{\bf ALOV~\cite{smeulders2013visual} and BDD100K~\cite{yu2020bdd100k}:} ALOV dataset is a diverse real-world video sequence for tracking, ranging from easy to difﬁcult, which covers as diverse circumstances as possible. This dataset consists of 315 video sequences and we capture 75,581 image pairs from them. BDD100K dataset is a large-scale diverse driving video database. It consists of 100,000 videos covering different weather conditions and urban scenes, ranging from daytime to nighttime. We sampled part of the videos and capture 86,128 images pairs. There is no flow label for these image pairs, so we use RealFlow to create a large diverse real-world dataset with flow label, named {\bf RF-AB}.
		
		\noindent{\bf KITTI~\cite{KITTI_2012,KITTI_2015}:} KITTI2012 and KITTI2015 dataset takes advantage of autonomous driving platform to develop challenging benchmarks for the tasks of optical flow estimation. They provide around $200$ training pairs and $200$ test pairs. There are multi-view extensions (4,000 training and 3,989 testing) datasets with no ground truth. We use the multi-view extension videos(training and testing) of KITTI 2015 to generate new datasets {\bf RF-Ktrain} and {\bf RF-Ktest}.
		
		\noindent{\bf Sintel~\cite{butler2012naturalistic}:} Sintel is a synthetic flow benchmark derived from the open-source 3D animated short film, which contains 1,041 training pairs and 564 test pairs. 'Clean' and 'Final' rendering sets are provided. We use the images from the training set to generate our {\bf RF-Sintel}.

		\section{Dataset Quantity}
		In this section, we analyze the impact of different amounts of datasets for network training. Specifically, we reduce the amount of our RF-AB dataset to half, one quarter, and one-sixteenth, respectively. Then we train RAFT from scratch on these datasets and evaluate the learned models on the train sets of KITTI 2015 and Sintel Final. 
		As shown in Table.~\ref{table:amount}, a better supervised network can be obtained by training on a larger dataset, which demonstrates that the quantity of labeled pairs is a crucial characteristic of optical flow dataset. 
		Moreover, we also compare our datasets with dCOCO~\cite{aleotti2021learning}. 
		We can notice that training on our RF-AB dataset with only $10,000$ pairs can achieve a better result than on dCOCO, which contains $100,000$ training pairs. 
		This phenomenon further proves that the realism of motion is a crucial factor for dataset generation. 
		
		\section{Splatting and Hole Filling}
		In our Realistic Image Pair Rendering (RIPR) method, we first forward-warp the reference image to the target view and then use splatting and our Bi-directional Hole Filling (BHF) to address the occlusion and hole problems. Here, we first present how the occlusion and hole problems occur in the forward warping process. Then, we show some visual results to analyze the details of the splatting and hole filling methods. 
		
\begin{figure}[t]
			\centering
			\includegraphics[width=0.6\linewidth]{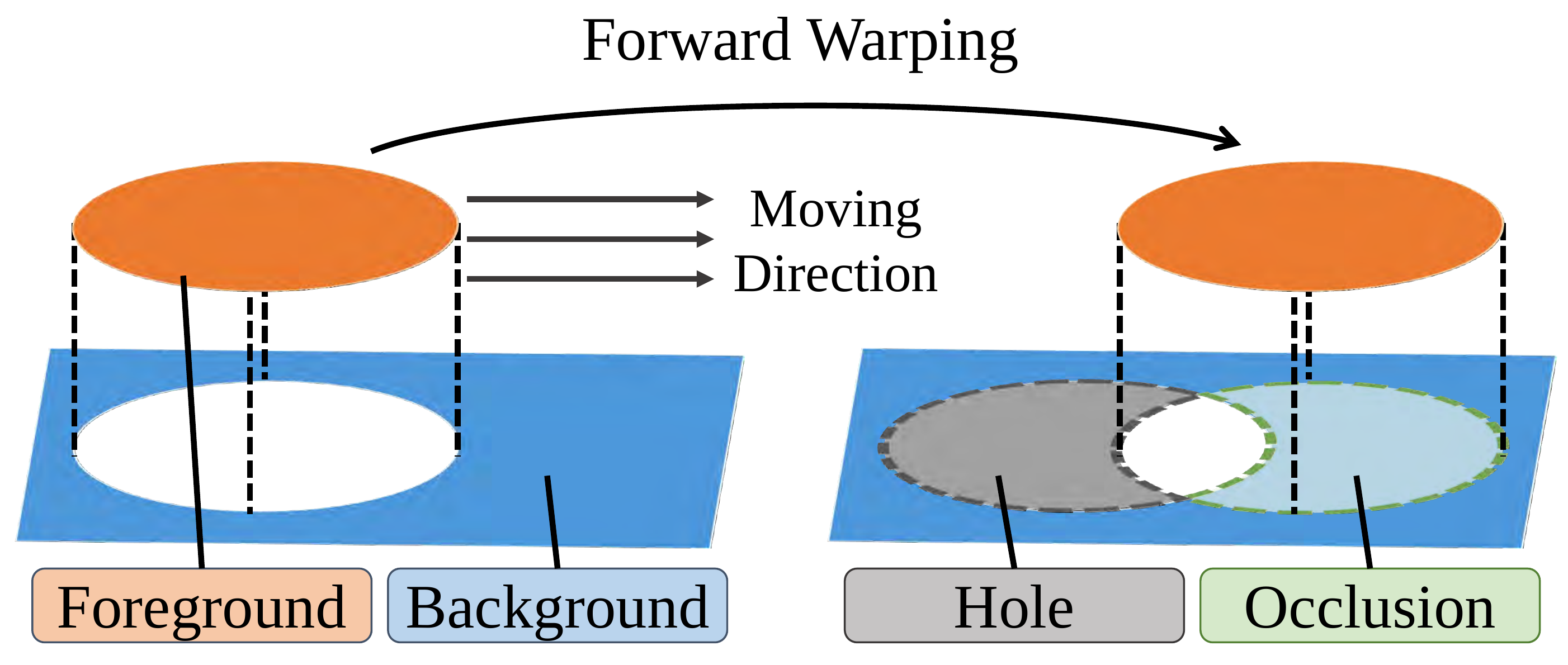}
			\caption{A toy example of how the hole and occlusion problems occur in the forward warping process. The reference image (left) is divided into two regions: the foreground object (the orange ellipse) and the background (the blue plane). The foreground object is moving to the right. When applying forward warping, the hole problem occurs in the gray region because no pixel is assigned to this area. The occlusion problem occurs in the green region because multiple pixels are assigned to this area. }
			\label{fig:splatting_and_hole_filling_explain}
		\end{figure}

		Fig.~\ref{fig:splatting_and_hole_filling_explain} is a toy example to illustrate how the hole and occlusion problems occur.
		We divide the reference image into the foreground region (the orange ellipse) and background region  (the blue plane), where the foreground region is moving to the right in the next view and the background remains stationary. 
		When the reference image is forward-warped to the next view, the foreground pixels and some background pixels are assigned to the same place, thus the occlusion problem occurs as the green region highlights.
		Besides, the hole problem occurs when no pixel is assigned to some areas as indicated by the gray region in Fig.~\ref{fig:splatting_and_hole_filling_explain}.
		
		\begin{figure*}[t]
			\centering
			\includegraphics[width=1.0\linewidth]{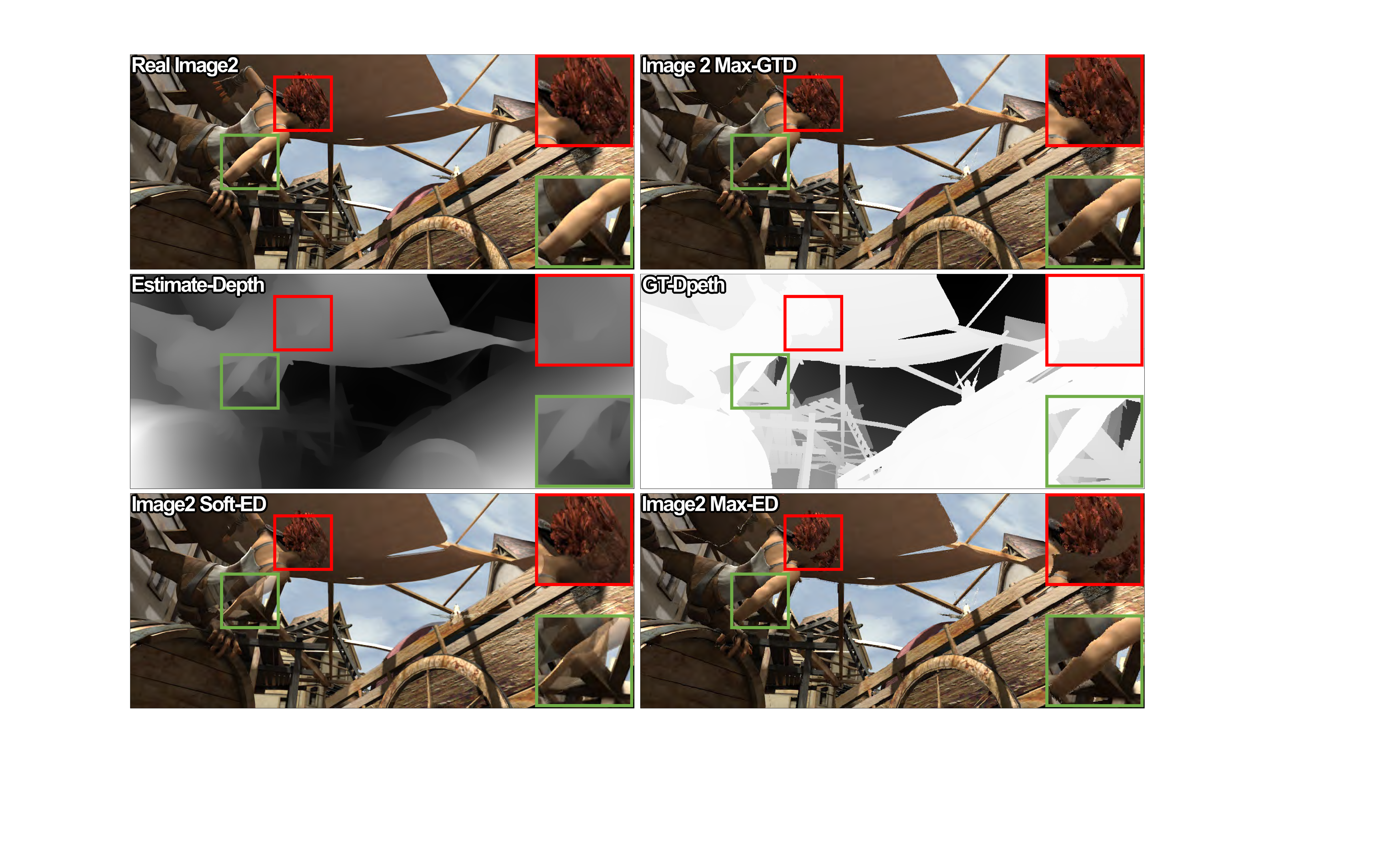}
			\caption{An example to show softmax splatting and max splatting with ground truth depth and the estimated depth by DPT~\cite{ranftl2021vision}. Zoom-in patches are shown in the right side. From top to bottom: the original image2 and the generated image2 by max splatting with ground truth depth, the estimated depth and ground truth depth, the softmax splatting result and max splatting result based on the estimated depth. We can notice that when the depth is correct, using max splatting can obtain realistic rendering result, as shown by the green boxes. However, when depth is incorrect, using max splatting may cause harmful tearing of the texture. This problem can be alleviated by using softmax splatting, as shown by the red boxes. }
			\label{fig:splatting_result}
		\end{figure*}

		\begin{figure*}[t]
			\centering
			\includegraphics[width=1.0\linewidth]{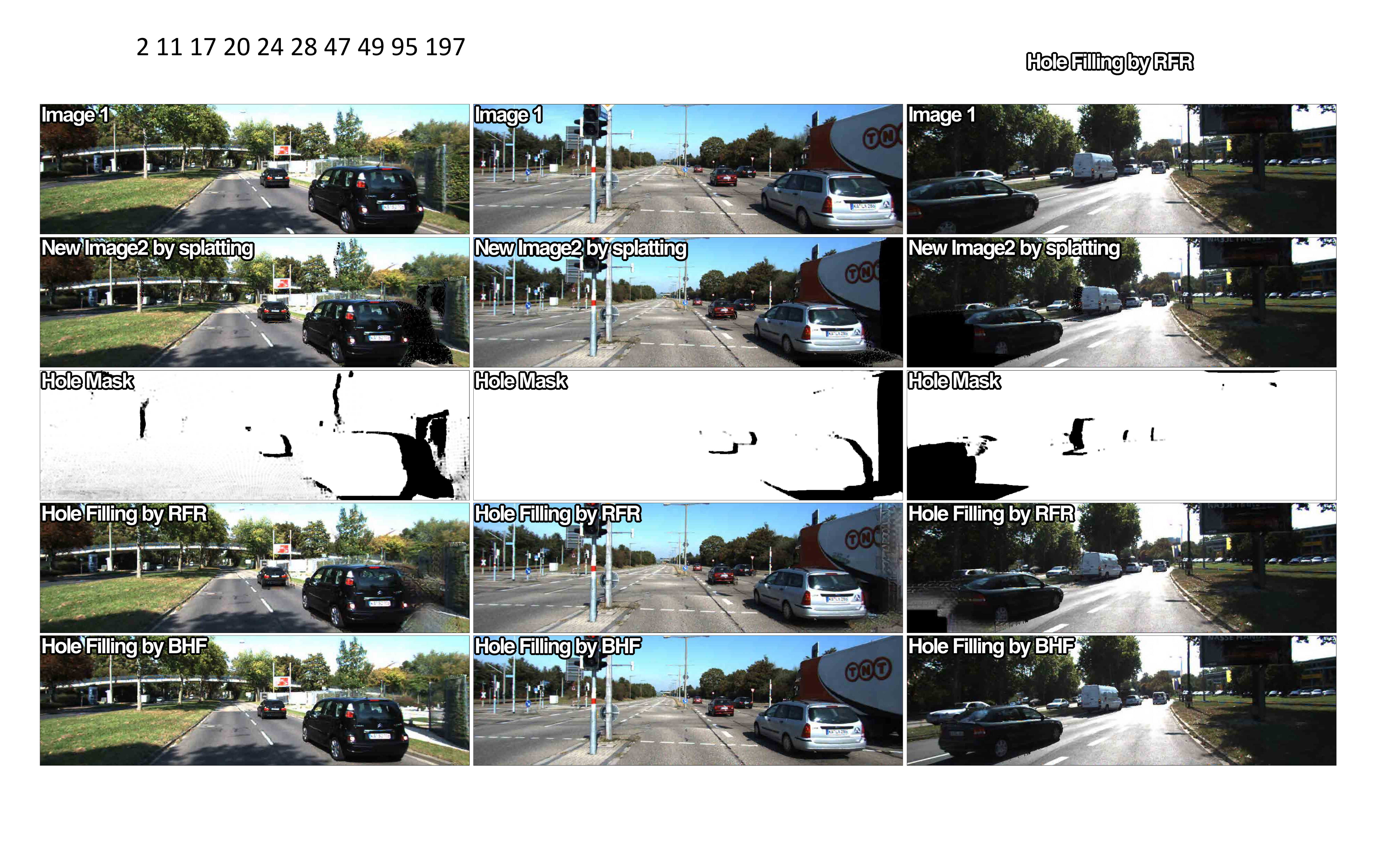}
			\caption{Qualitative comparison of different methods for hole filling. Top to bottom: the reference images, the `new image 2' generated by splatting, the hole masks, hole filling results by RFR~\cite{li2020recurrent}, and hole filling results by our BHF method. }
			\label{fig:hole-fill}
		\end{figure*}
		
		\par For splatting, as mentioned in our main text, max splatting renders more realistic images than softmax splatting. However, we notice that softmax splatting can produce better dataset than max splatting in our ablation study. Here, we explain why this happens by an example from Sintel dataset that is shown in Fig.~\ref{fig:splatting_result}. 
		We use softmax splatting and max splatting to warp the reference image to the target view based on ground truth optical flow and different depth maps. 
		In Fig.~\ref{fig:splatting_result}, the first line shows the original target image `Real Image2' and the generated target image `Image2 Max-GTD' by using max splatting based on the ground truth depth. The second line shows the estimated depth by DPT~\cite{ranftl2021vision} and the ground truth depth. The third line shows the softmax splatting result and max splatting result based on the estimated depth.
		As can be seen from Fig.~\ref{fig:splatting_result}, when the depth map is correct (using ground truth depth or correct depth estimation), max splatting method can generate a realistic result, which is almost the same as the original target image. 
		However, as highlighted by the red box, the incorrect depth may cause harmful tearing of the texture in the max splatting result, which may reduce the quality of the dataset generated by the max splatting method. 
		This problem can be alleviated by using softmax splatting method, which can obtain a translucent fusion result in these occlusion areas so that the information of the foreground objects can be partially provided even with incorrect depth estimation. This is the reason why sofmax splatting can produce better results than the max version, though max splatting can provide image pairs look more realistic.
		
		\par For hole filling, in ablation study, we have validated that our BHF outperforms the deep neural inpainting method RFR~\cite{li2020recurrent}. Fig.~\ref{fig:hole-fill} presents some examples for qualitative comparison. For the fair comparison, we use the RFR~\cite{li2020recurrent} pre-trained on the large-scale outdoor dataset PARIS, and fine-tuned on KITTI dataset with 100k steps to inpaint these holes. As can be seen, inpainting method RFR may introduce artifacts in these hole regions, while our BHF method can produce realistic image for dataset construction. 
		\begin{figure*}
			\centering
			\includegraphics[width=0.6\linewidth]{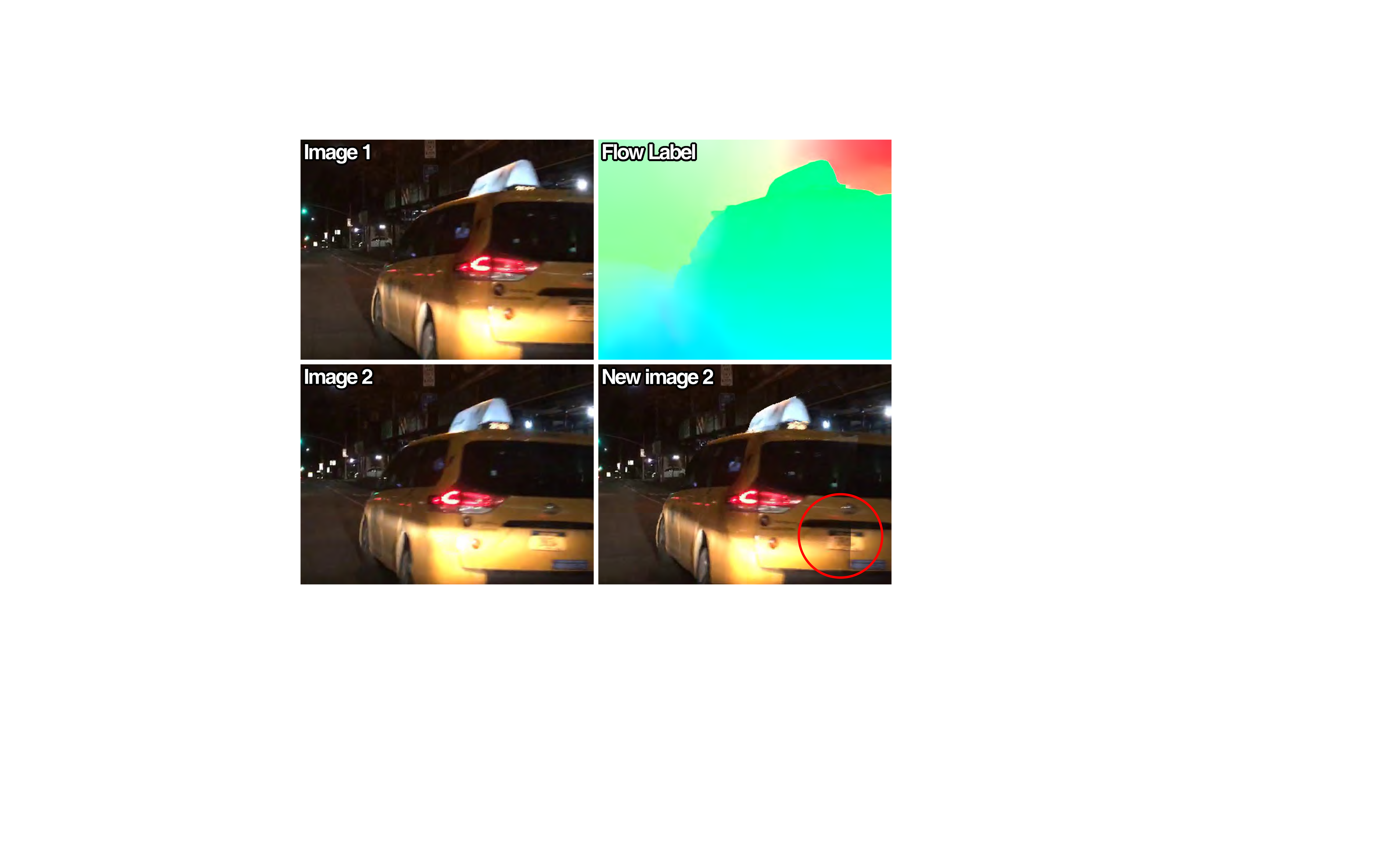}
			\caption{Limitation of our RealFlow. As pointed by the red circle, there is discontinuous illumination artifact in the generated new image 2. This is because that the brightness of the car changes greatly in the original image pair.  }
			\label{fig:limitation}
		\end{figure*}
		
		\begin{figure*}[t]
			\centering
			\includegraphics[width=1.0\linewidth]{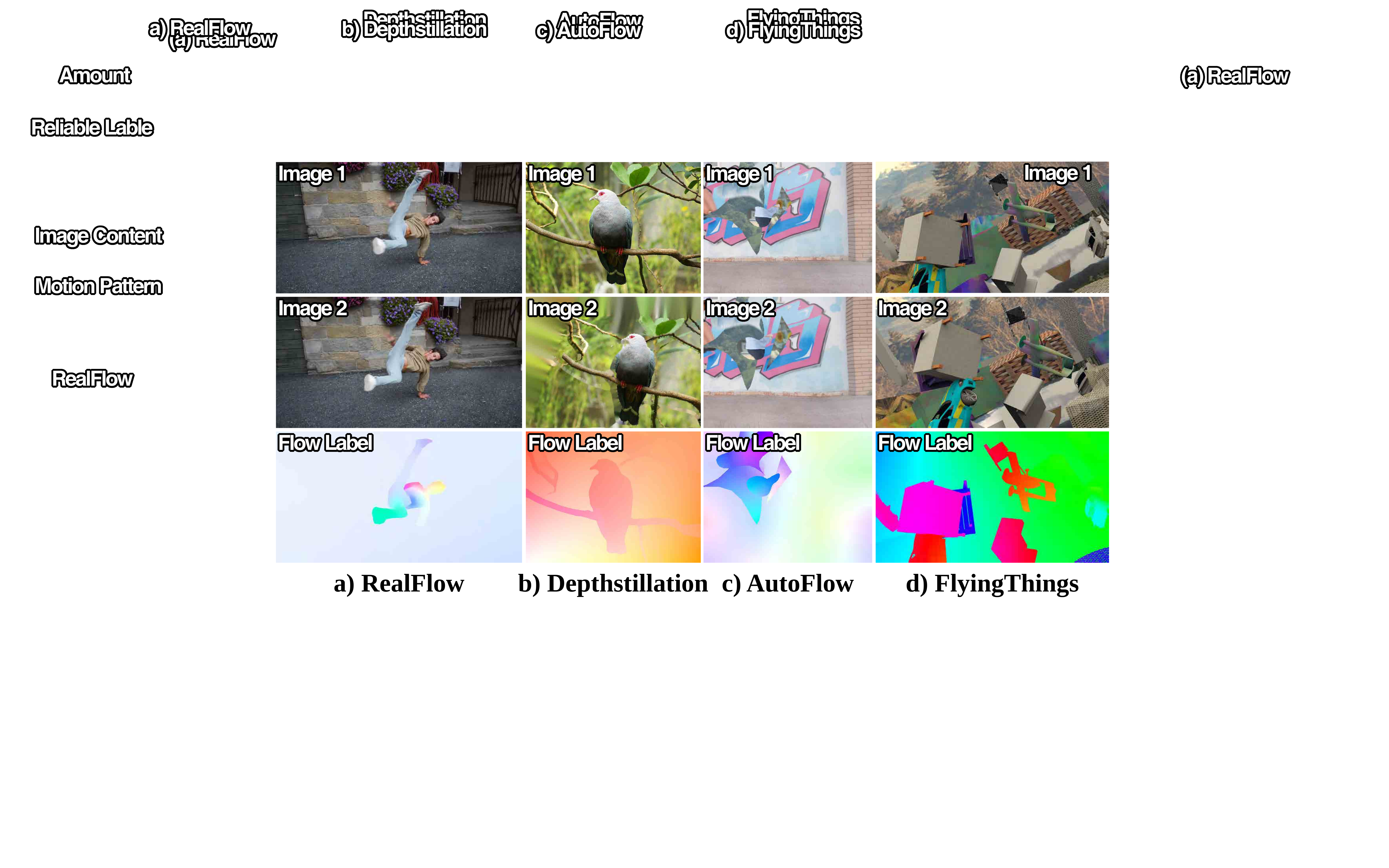}
			\caption{Example training pairs of our RealFlow and other dataset generation methods, such as Depthstillation~\cite{aleotti2021learning}, AutoFlow~\cite{sun2021autoflow} and FlyingThings~\cite{yang2019drivingstereo}. }
			\label{fig:dataset_sample_comparison}
		\end{figure*}
		
		\section{Efficiency of RealFlow}
		The efficiency is an essential indicator of a dataset generation method. 
		Traditional manually annotation method~\cite{liu2008human} may cost $20\sim30$ minutes per image pair, especially for challenging scenes.
		AutoFlow\cite{sun2021autoflow} takes about 7 days to finish 8 searching iterations using 48 NVIDIA P100 GPUs for the rendering hyperparameters searching and then use the learned hyperparamters to render image pairs.
		In contrast, our method can automatically generate a large amount of training pairs with high efficiency. 
		During the E-step, the running time of our RealFlow is 0.53s for generating a training pair with a resolution of $512\times 960$ using only 1 NVIDIA 2080Ti GPU. The time consumption of the M-step in our RealFlow depends on the training of the deep neural network. In this work, we train RAFT~\cite{raft2020} for 120k iterations with a batch size of 5 on 1 NVIDIA 2080Ti GPU. Using more computational resources may reduce the time consumption of our RealFlow.
		
		\section{Limitation}
		In Fig.~\ref{fig:limitation}, as illustrated by the red circle, there is discontinuous illumination artifact in the generated new image 2. This is because that the brightness of the car changes greatly in the original image pair. During the hole filling process of our RealFlow, object regions with different brightness are used to fill the occlusion regions. 
		
		\section{Qualitative Results}
		
		\subsection{Comparison with Dataset Generation Methods}
		\par In Fig.~\ref{fig:dataset_sample_comparison}, we show example training pairs of our RealFlow, Depthstillation~\cite{aleotti2021learning}, AutoFlow~\cite{sun2021autoflow} and FlyingThings~\cite{yang2019drivingstereo}. As can be seen, the training pair of Depthstillation contains a large region of artifacts that reduce the realism of the image. AutoFlow and FlyingThings samples are generated by moving the foreground image patches or objects on the background image, thus the scene objects and their motions are synthesized and can not match the real-world scene.
		Compared with these methods, the motion and image content of our RealFlow is realistic because the training pair is generated from real-world videos.
		
		\begin{figure*}[t]
			\centering
			\includegraphics[width=0.6\linewidth]{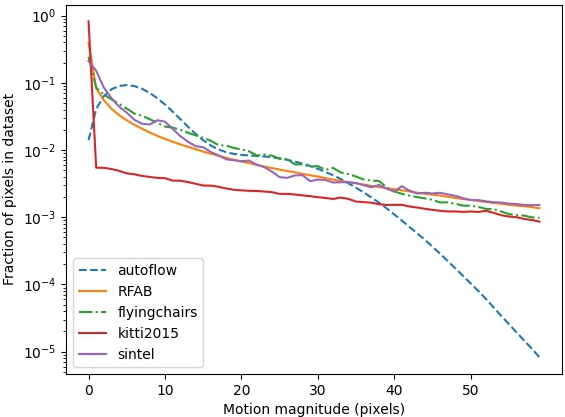}
			\caption{Histogram of motion magnitude for the generated dataset RFAB and existing datasets.}
			\label{fig:motion magnitude}
		\end{figure*}
		
	
			\begin{figure*}[t]
		\centering
		\subfigure[Qualitative comparison of RAFT trained on our RF-DAVIS dataset with that trained on dDAVIS~\cite{aleotti2021learning} dataset.]{
			\includegraphics[width=1.0\linewidth]{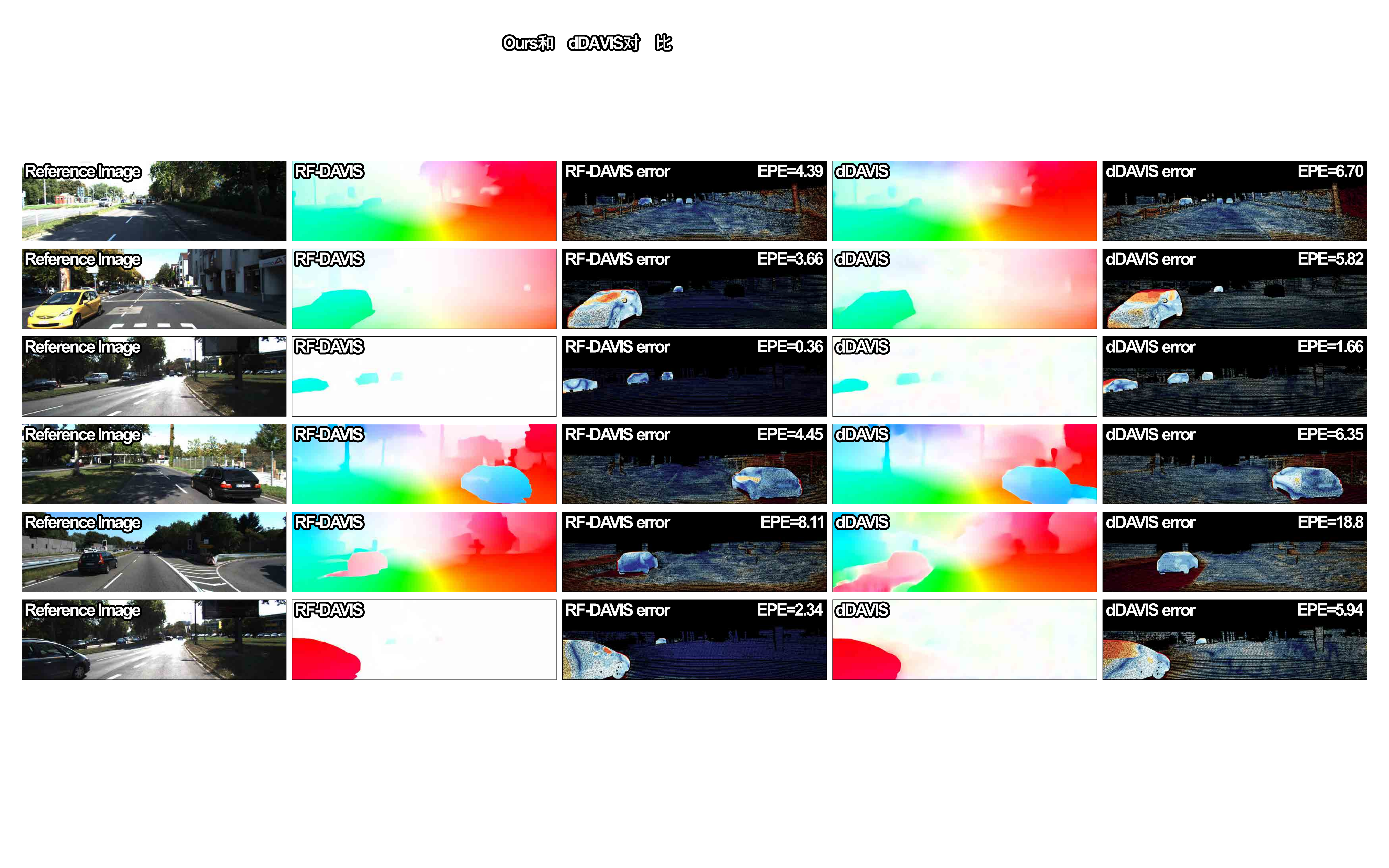}
			\label{fig:ours_with_depthstll_DAVIS}}\\
		\centering
		\subfigure[Qualitative comparison of RAFT trained our RF-Ktest dataset with that trained on dKITTI~\cite{aleotti2021learning} dataset.]{
			\includegraphics[width=1.0\linewidth]{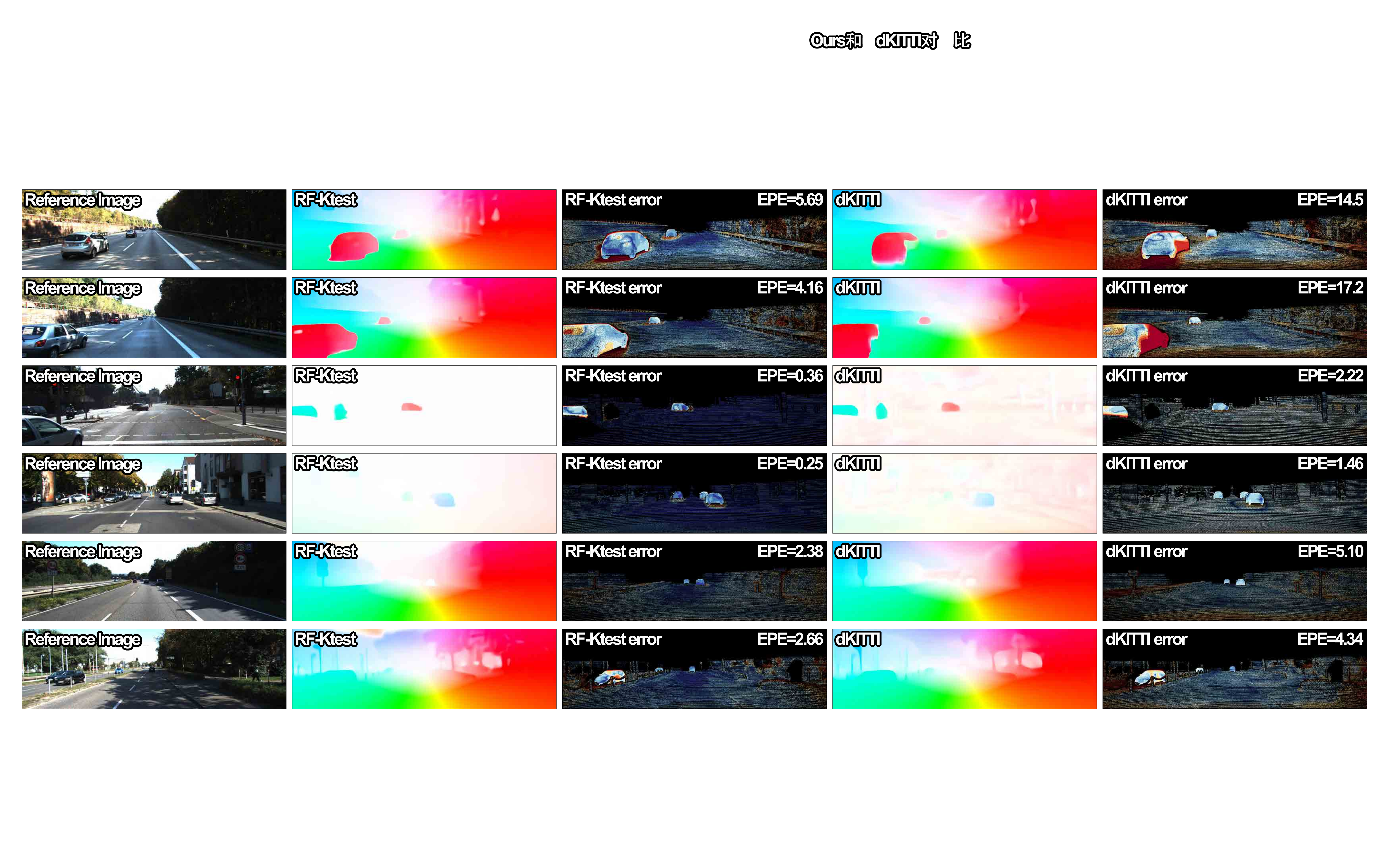}
			\label{fig:teaser_b_upsample_vis}}
		\caption{Qualitative comparison of our RealFlow with previous dataset generation method Depthstillation~\cite{aleotti2021learning}. Depthstillation generated optical flow dataset dDAVIS and dKITTI from DAVIS and KITTI multiview test. We also use the DAVIS and KITTI multi-view test videos to generate our RF-DAVIS and RF-Ktest datasets. We train the same model RAFT on these datasets and evaluate on KITTI 2015 train set. Error maps are also visualized, where correct pixels are displayed in blue and wrong ones in red.
		}\label{fig:ours_with_depthstll}
	\end{figure*}
		
		The statistics of motion magnitude for different datasets have been shown in Fig.~\ref{fig:motion magnitude}. Our RF-AB dataset exhibits an exponential falloff like Sintel and Flyingchairs but is more smooth. The motion of KITTI mainly concentrates in small range which forms a steep polyline. AutoFlow has few small motions and focuses on middle-range motions. The statistics of motion magnitude for RF-AB is similar to the KITTI, when compared with flyingchairs and Autoflow, which maybe one of the factors that lead to the effectiveness when evaluating on KITTI.
		
		\par We also provide qualitative comparison of our RealFlow with previous dataset generation method Depthstillation~\cite{aleotti2021learning} in Fig.~\ref{fig:ours_with_depthstll}. Specifically, Depthstillation uses DAVIS and KITTI multi-view test videos to generate optical flow training dataset dDAVIS and dKITTI. For fair comparison, we also use the same source videos to generate RF-DAVIS and RF-Ktest datasets. Then we train RAFT on these datasets with the same hyperparameters and do evaluation on KITTI 2015 train set. The error map of each sample is also visualized with the EPE error depicted in the top-right corner. 
		As can be seen from Fig.~\ref{fig:ours_with_depthstll}, models trained on our RF-DAVIS and RF-Ktest datasets can produce better optical flow estimation results than trained on dDAVIS and dKITTI, which demonstrates the effectiveness of our dataset generation method RealFlow. 
		
		\subsection{Comparison with Unsupervised Methods}
\begin{figure*}
			\centering
			\includegraphics[width=1.0\linewidth]{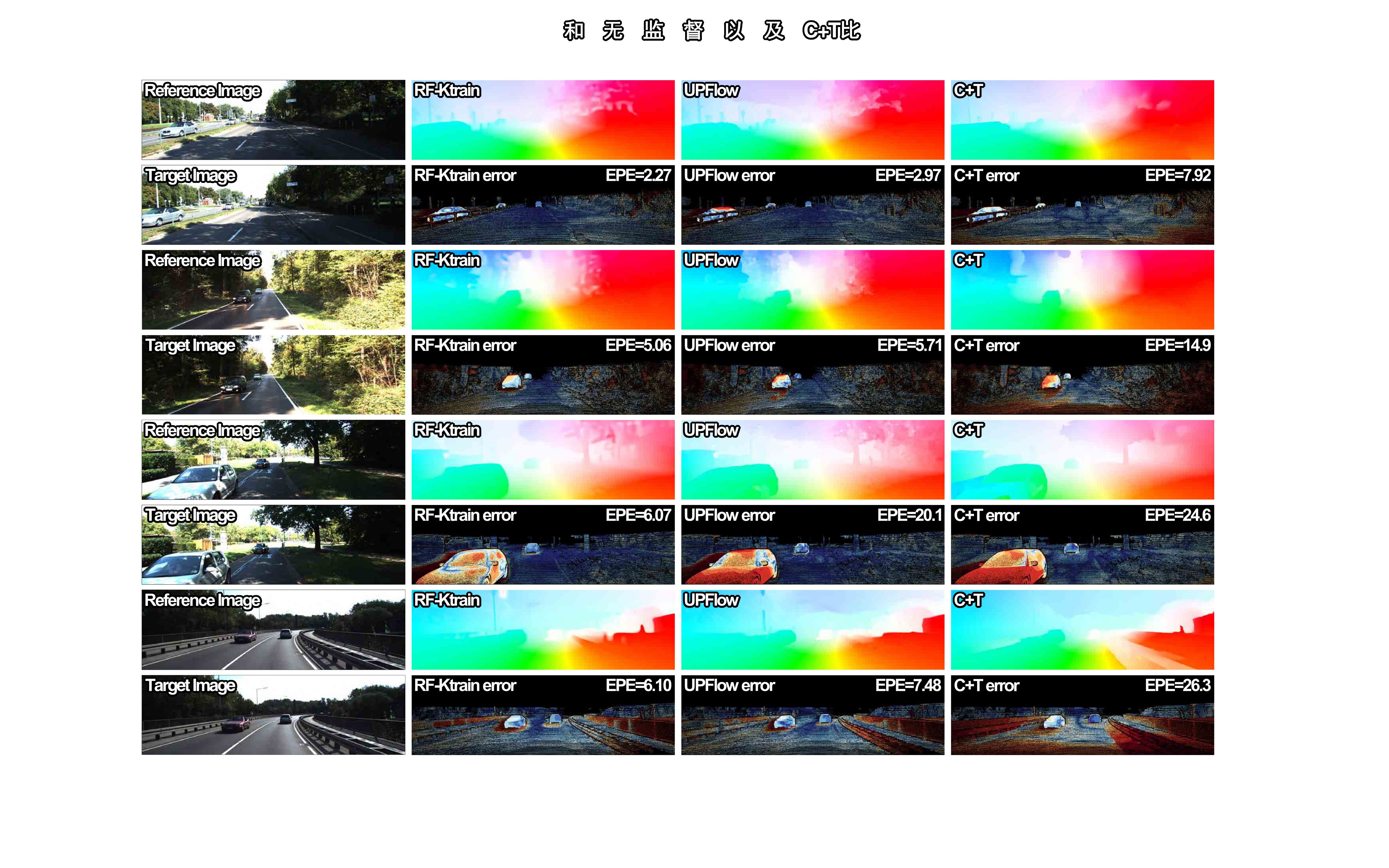}
			\caption{Qualitative comparison of our RealFlow with unsupervised methods. 
				We first generate RF-Ktrain dataset using the same training videos as the unsupervised method UPFlow~\cite{luo2021upflow}. Then we train RAFT on our RF-Ktrain to obtain the optical flow predictions for comparison with the unsupervised method UPFlow and the baseline method, where RAFT is trained on C+T (synthetic dataset FlyingChairs and FlyingThings). Error maps are visualized, where correct predictions are displayed in blue and wrong ones in red. The End-Point-Error (epe) is used as the evaluation metric, which is also depicted in the top-right side of each sample.}
			\label{fig:ours_with_unsupervised}
		\end{figure*}

		There is a set of unsupervised methods that can learn deep neural networks using only video sequences without optical flow labels. We also compare our RealFlow with state-of-the-art unsupervised methods. We first generate an optical flow dataset RF-Ktrain using KITTI multi-view train set, which is also used as the training set of the unsupervised methods. Then we train RAFT on our RF-Ktrain and do evaluation on KITTI 2015 train set. We show qualitative comparison result in Fig.~\ref{fig:ours_with_unsupervised}, where `C+T' is the baseline model trained on synthetic datasets FlyingChairs and FlyingThings, UPFlow~\cite{luo2021upflow} is the state-of-the-art unsupervised method that trained on the same source videos mentioned above. The End-Point-Error (EPE) is used as the evaluation metric and the error maps are also visualized, where correct predictions are depicted in blue and wrong ones in red. As can be seen, optical estimation network trained on our generated dataset can produce better results than unsupervised methods. 
		
		\subsection{Comparison with Supervised Methods}
\begin{figure*}
			\centering
			\includegraphics[width=1.0\linewidth]{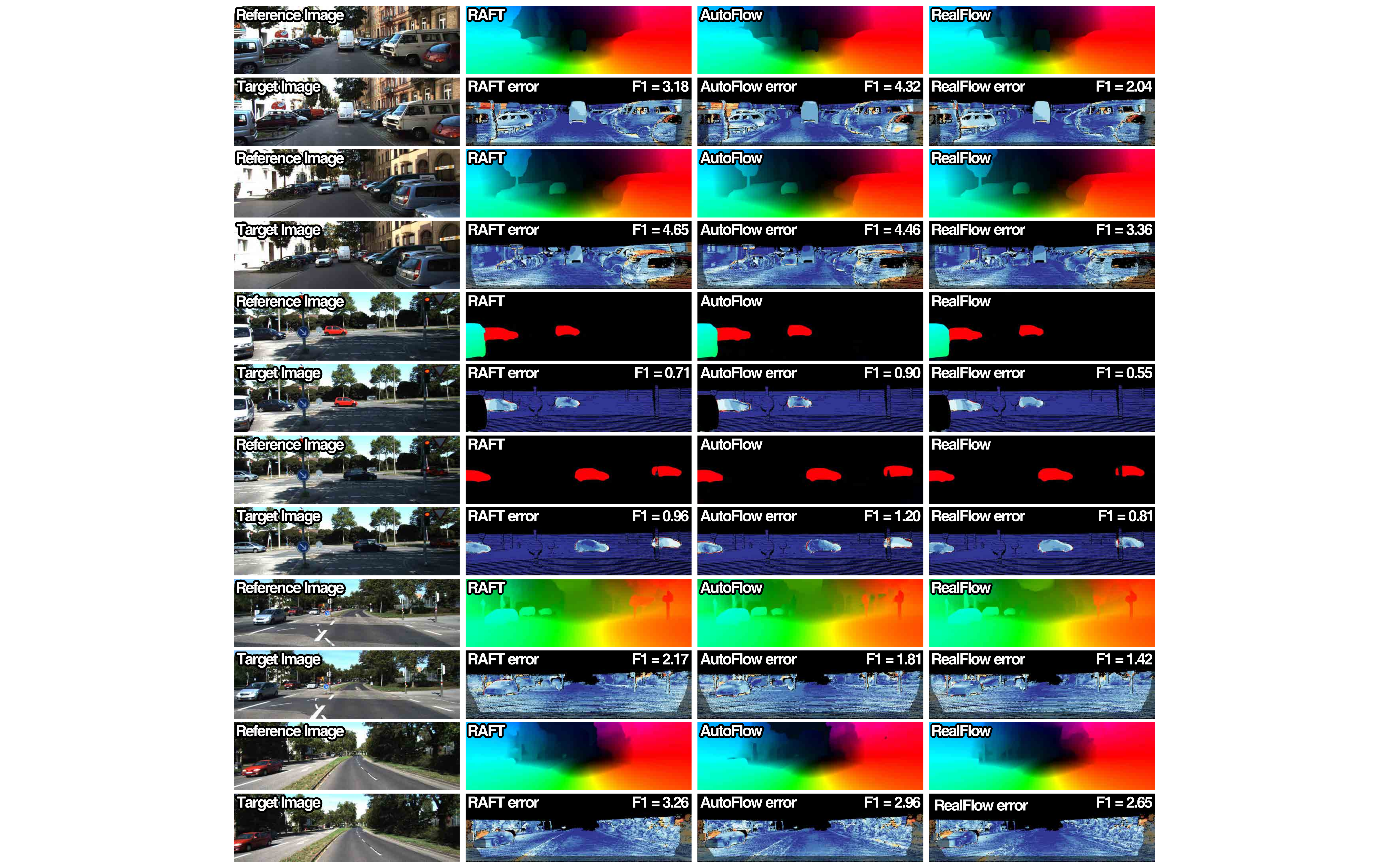}
			\caption{Qualitative comparison of our method with AutoFlow and RAFT on KITTI 2015 online benchmark. 
				The percentage of erroneous pixels (F1) is used as the evaluation metric. 
				Error maps are visualized by KITTI 2015 website, where correct pixels are displayed in blue and wrong ones in red.}
			\label{fig:ours_with_supervised}
		\end{figure*}

		The original RAFT~\cite{raft2020} is first pre-trained on synthetic datasets FlyingChairs and FlyingThings and then fine-tuned on KITTI15-training dataset. AutoFlow~\cite{sun2021autoflow} pre-train RAFT on their proposed AutoFlow dataset and follow the same procedure to further fine-tune the model. We also pre-train the RAFT network using our RF-train dataset generated by RealFlow and then fine-tune on KITTI15-training dataset. We compare our results with them on the public online benchmark KITTI 2015 testing. The percentage of erroneous pixels (F1) is used as the evaluation metric. In Fig.~\ref{fig:ours_with_supervised}, we show some qualitative comparison results of our method with AutoFlow and the original RAFT.
		As can be seen, our method achieves better results than AutoFlow and RAFT.
		
		\subsection{Visualization of RealFlow samples}
		Our RealFlow can automatically generate training pairs from videos without human involvement. Thus, huge amount of videos can be used to generate optical flow training pairs which help the supervised networks generalize to various scenes.
		\par
\begin{figure*}
			\centering
			\includegraphics[width=1.0\linewidth]{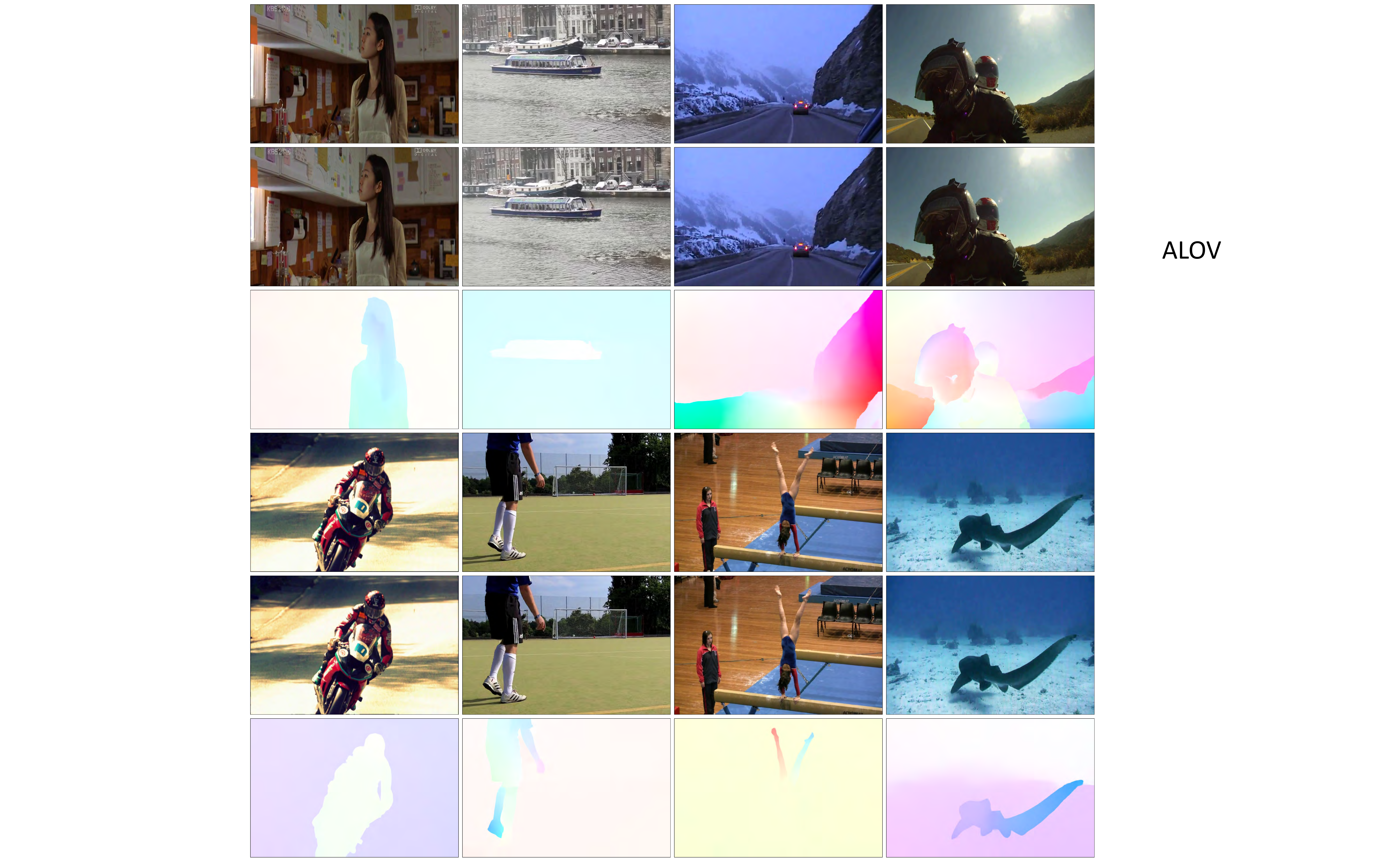}
			\caption{Samples from RF-AB. These samples are generated from ALOV~\cite{smeulders2013visual} which is a large-scale real-world video dataset that covers various circumstances. First/fourth row: image 1; second/fifth row: generated new image 2; third/sixth row: visualized optical flow. Note that, these flow maps are the ground truth of our generated pairs.}
			\label{fig:ALOV_samples}
		\end{figure*}

\begin{figure*}
			\centering
			\includegraphics[width=1.0\linewidth]{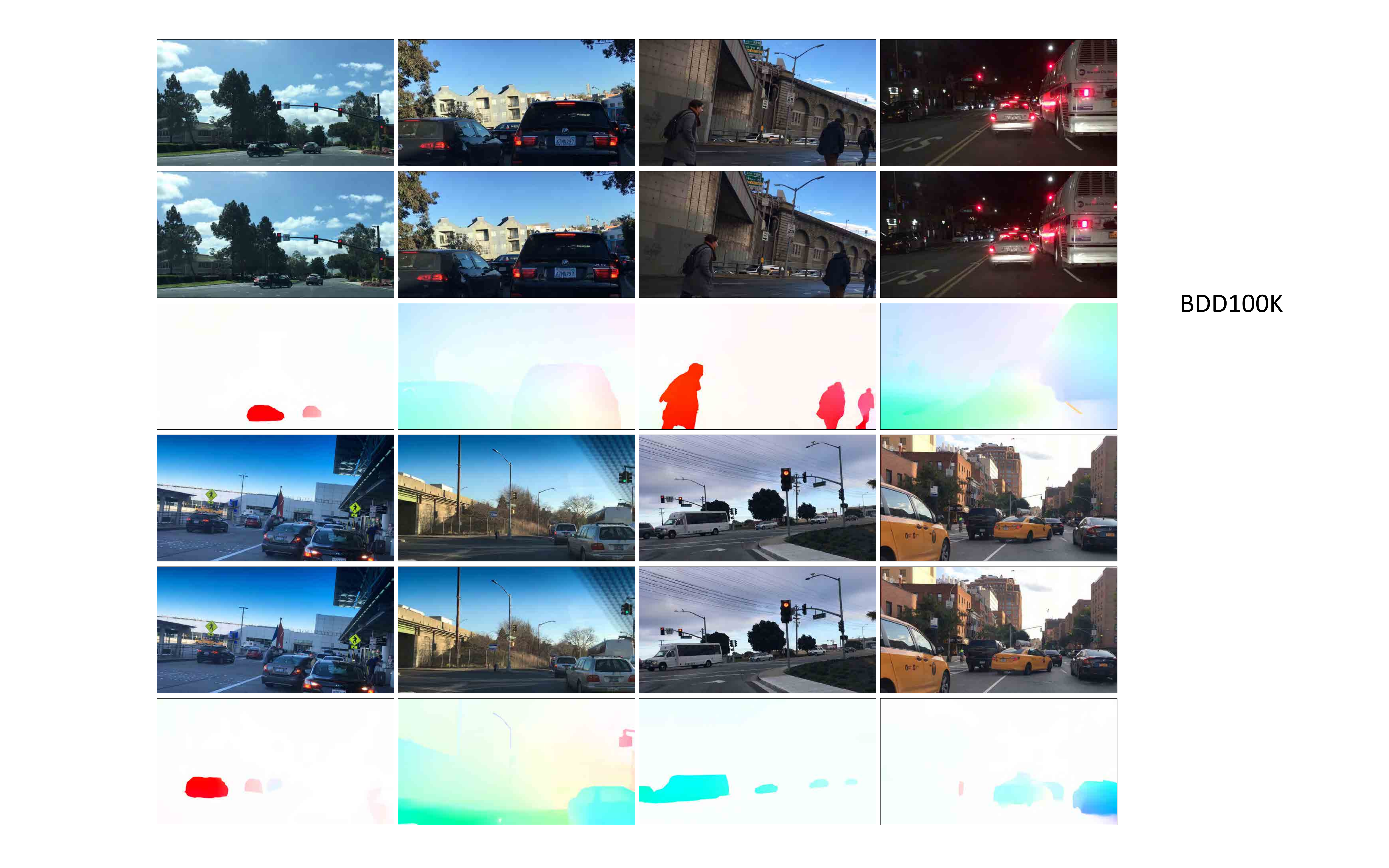}
			\caption{Samples from RF-AB. These samples are generated from BDD100K~\cite{yu2020bdd100k} which is a large-scale diverse driving video database covering various urban scenes. First/fourth row: image 1; second/fifth row: generated new image 2; third/sixth row: visualized optical flow. Note that, these flow maps are the ground truth of our generated pairs. }
			\label{fig:BDD100K_samples}
		\end{figure*}

 Fig.~\ref{fig:ALOV_samples} shows some samples generated from ALOV~\cite{smeulders2013visual} dataset. ALOV dataset contains diverse real-world video sequences that cover various circumstances.
		Fig.~\ref{fig:BDD100K_samples} shows samples generated from BDD100K~\cite{yu2020bdd100k} dataset. BDD100K is a large-scale diverse driving video database that covers diverse urban scenes. 
		The samples generated from ALOV and BDD100K are used to construct our dataset RF-AB. \par
\begin{figure*}
			\centering
			\includegraphics[width=1.0\linewidth]{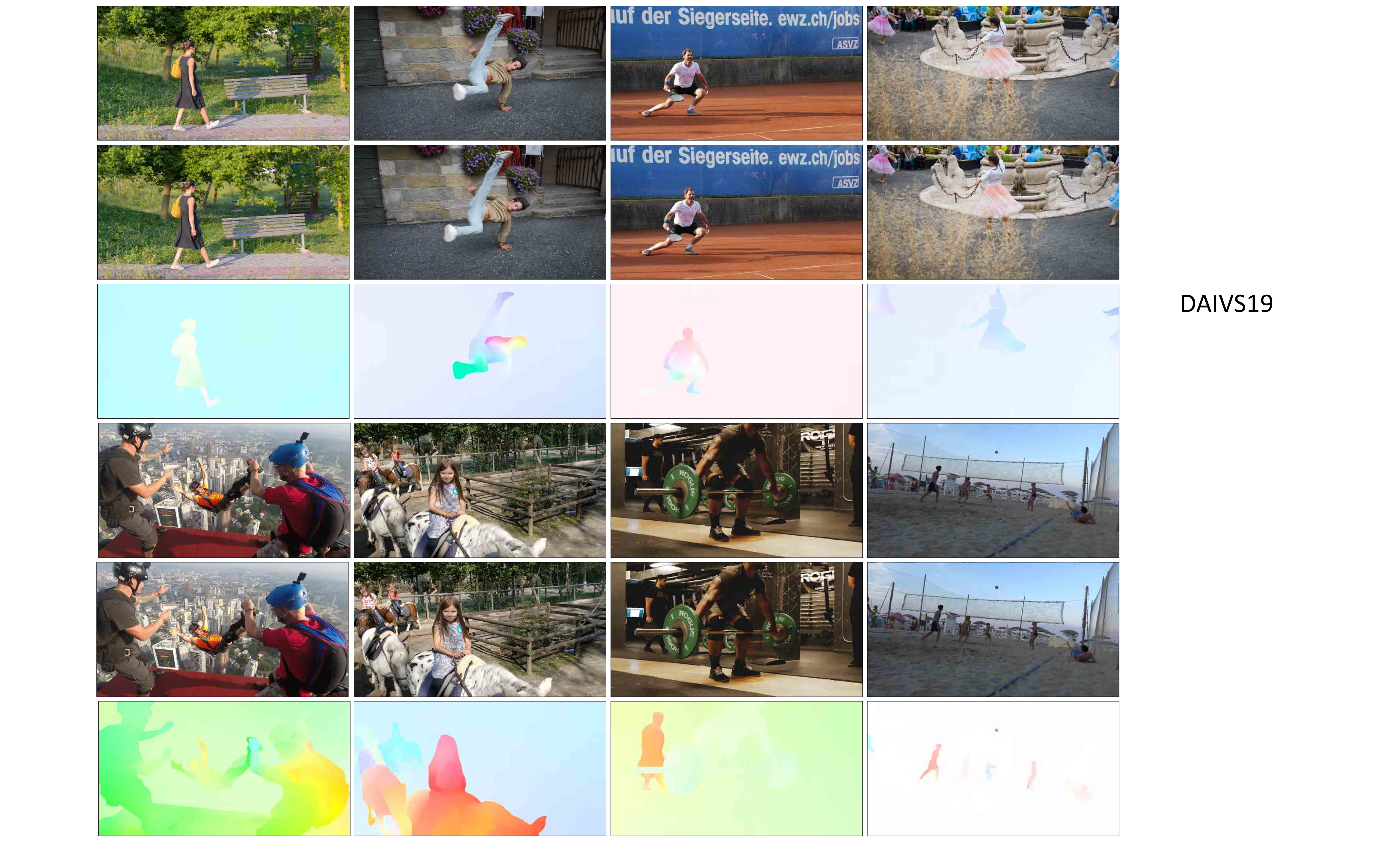}
			\caption{Samples from RF-DAVIS. These samples are generated from DAVIS challenge 2019~\cite{Caelles_arXiv_2019} which consists of high-quality video sequences under various kinds of scenes. First/fourth row: image 1; second/fifth row: generated new image 2; third/sixth row: visualized optical flow. Note that, these flow maps are the ground truth of our generated pairs. }
			\label{fig:DAVIS_samples}
		\end{figure*}

 Fig.~\ref{fig:DAVIS_samples} shows some samples generated from DAVIS~\cite{Caelles_arXiv_2019} challenge 2019. DAVIS consists of high-quality video sequences under various kinds of scenes. The samples generated from DAVIS are used to construct our dataset RF-DAVIS. 
		\par
\begin{figure*}
			\centering
			\includegraphics[width=1.0\linewidth]{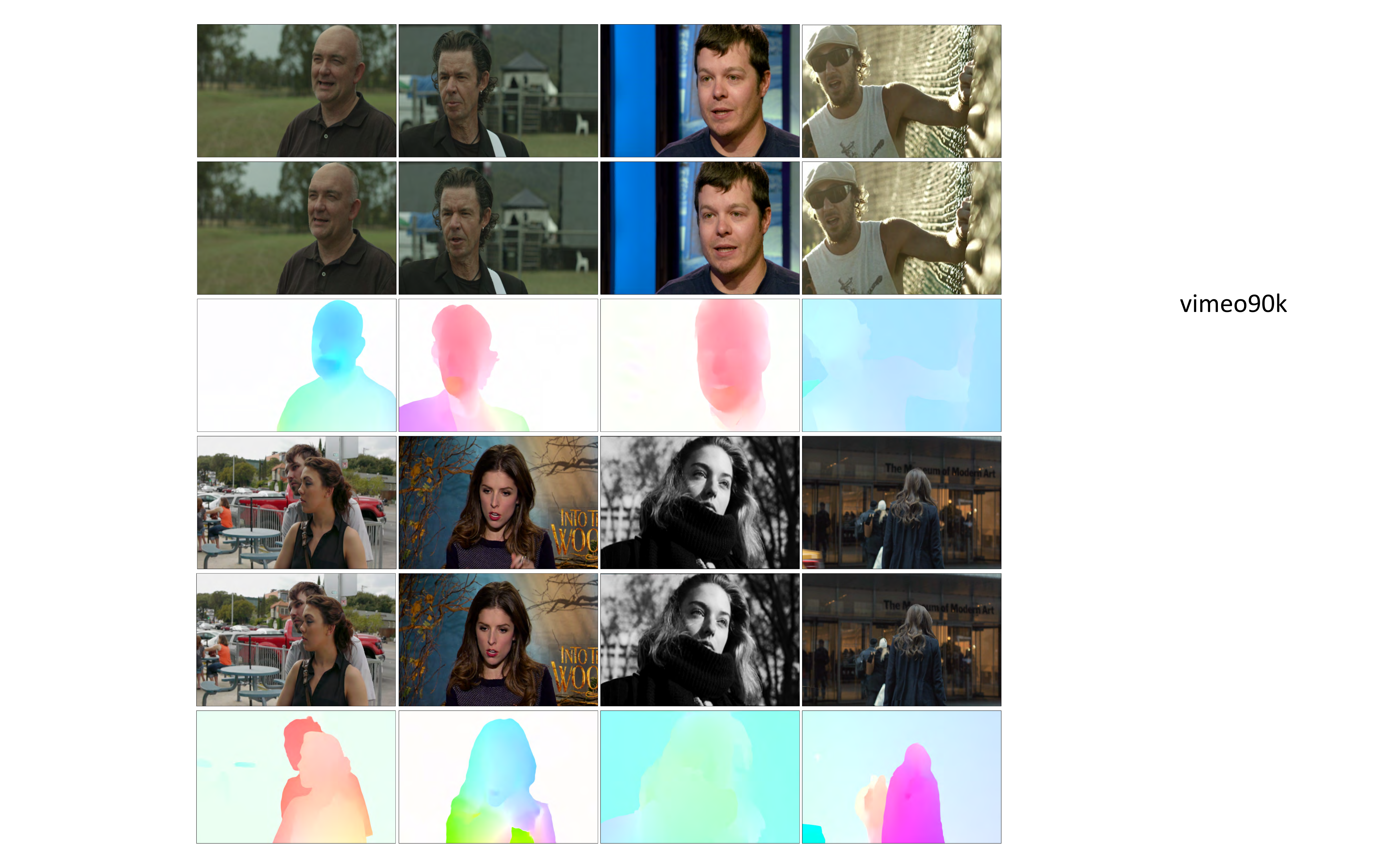}
			\caption{Vimeo-90k Dataset~\cite{xue2019video} is a large-scale, high-quality video dataset collected for video enhancement tasks such as temporal frame-interpolation, denoising and super-resolution. We also generated some training pairs from Vimeo-90k which may be helpful for some specific scenes. First/fourth row: image 1; second/fifth row: generated new image 2; third/sixth row: visualized optical flow. Note that, these flow maps are the ground truth of our generated pairs.}
			\label{fig:vimeo90k_samples}
		\end{figure*}

 Besides, in order to demonstrate the efficiency and versatility of our method, we also use Vimeo-90k~\cite{xue2019video} dataset to generate some samples which is shown in Fig.~\ref{fig:vimeo90k_samples}. Vimeo-90k Dataset is a large-scale, high-quality video dataset for video enhancement such as temporal frame-interpolation, denoising, and super-resolution, which contains many close-ups. These samples may be helpful for some specific tasks.
		
	\end{appendices}

\end{document}